\documentclass[10pt,times,mathptm,psfig,twocolumn,journals]{IEEEtran}
\usepackage{url}
\usepackage{amssymb}
\usepackage{array}
\usepackage{amsmath}
\usepackage{graphicx}
\usepackage{booktabs}
\usepackage{epsfig}
\usepackage{ulem}
\usepackage{hyperref}
\hypersetup{pdfborder={0 0 0}, colorlinks=true, linkcolor=red, citecolor=blue}
\usepackage{subfigure}
\usepackage{cite}
\usepackage{epstopdf}
\usepackage{amsmath}
\usepackage[misc]{ifsym}
\usepackage{bm}
\usepackage{mathrsfs}
\usepackage{float}
\usepackage{setspace}
\usepackage{multirow}
\usepackage{tabularx}
\usepackage{url}
\usepackage{booktabs}
\usepackage{makecell}
\usepackage{epigraph}
\usepackage{bbding}
\usepackage{txfonts}

\begin{document}

\title{Depth-Aware Multi-Grid Deep Homography Estimation with Contextual Correlation}

\author{Lang~Nie, Chunyu~Lin,~\IEEEmembership{Member,~IEEE}, Kang~Liao,~\IEEEmembership{Student Member,~IEEE}, Shuaicheng Liu,~\IEEEmembership{Member,~IEEE}, Yao~Zhao,~\IEEEmembership{Senior Member,~IEEE}
\thanks{This work was supported by the National Natural Science
Foundation of China (No.62172032).
\textit{(Corresponding author: Chunyu Lin)}}
\thanks{Lang Nie, Chunyu Lin, Kang Liao, Yao Zhao are with the Institute of Information Science, Beijing Jiaotong University, Beijing 100044, China, and also with the Beijing Key Laboratory of Advanced Information Science and Network Technology, Beijing 100044, China (email: nielang@bjtu.edu.cn, cylin@bjtu.edu.cn, kang\_liao@bjtu.edu.cn, yzhao@bjtu.edu.cn).}
\thanks{Shuaicheng Liu is with School of Information and Communication Engineering, University of Electronic Science and Technology of China, Chengdu, 611731, China (liushuaicheng@uestc.edu.cn).}
}

\maketitle
%\graphicspath{{img/}}
\begin{abstract}
    Homography estimation is an important task in computer vision applications, such as image stitching, video stabilization, and camera calibration. Traditional homography estimation methods heavily depend on the quantity and distribution of feature correspondences, leading to poor robustness in low-texture scenes. The learning solutions, on the contrary, try to learn robust deep features but demonstrate unsatisfying performance in the scenes with low overlap rates. In this paper, we address these two problems simultaneously by designing a contextual correlation layer (CCL). The CCL can efficiently capture the long-range correlation within feature maps and can be flexibly used in a learning framework. In addition, considering that a single homography can not represent the complex spatial transformation in depth-varying images with parallax, we propose to predict multi-grid homography from global to local. Moreover, we equip our network with a depth perception capability, by introducing a novel depth-aware shape-preserved loss. Extensive experiments demonstrate the superiority of our method over state-of-the-art solutions in the synthetic benchmark dataset and real-world dataset. The codes and models will be available at \url{https://github.com/nie-lang/Multi-Grid-Deep-Homography}.
\end{abstract}
\begin{IEEEkeywords}
    Homography estimation, mesh deformation
\end{IEEEkeywords}

%\markboth{IEEE TRANSACTIONS ON IMAGE PROCESSING}
\markboth{}
{Shell \MakeLowercase{\textit{et al.}}: Bare Demo of IEEEtran.cls for IEEE Transactions on Magnetics Journals}
\IEEEpeerreviewmaketitle

\section{Introduction}
\label{section1}
A homography is an invertible mapping from one image plane to another with 8-DOF, including 2 for translation, 2 for rotation, 2 for scale, and 2 for lines at infinity. It has been widely served as an essential component of various computer vision applications, such as image stitching\cite{nie2020view, zaragoza2013projective}, video stabilization\cite{7331626, 9272336, liu2016meshflow}, camera calibration\cite{zhang2000flexible}, and simultaneous localization and mapping (SLAM)\cite{engel2014lsd}.

Traditional algorithms estimate the homography in two ways. Pixel-based approaches update the homography parameters by optimizing the pixel-level alignment error between two images in an iterative way \cite{10.5555/1623264.1623280}, often failing to deal with scenes with low overlap rates. Feature-based approaches search for the optimal homography model based on  sparse feature correspondences using different feature extractors\cite{lowe2004distinctive, rublee2011orb, leutenegger2011brisk, sosnet2019cvpr} and different robust estimation strategies\cite{fischler1981random, barath2019magsac, barath2019magsacplusplus}. However, they heavily depend on the number and distribution of feature correspondences, leading to poor robustness in low-texture scenes.

\begin{figure}[!t]
    \centering
    \subfigure[A scene of low overlap rate.]
    {\includegraphics[width=0.48\textwidth]{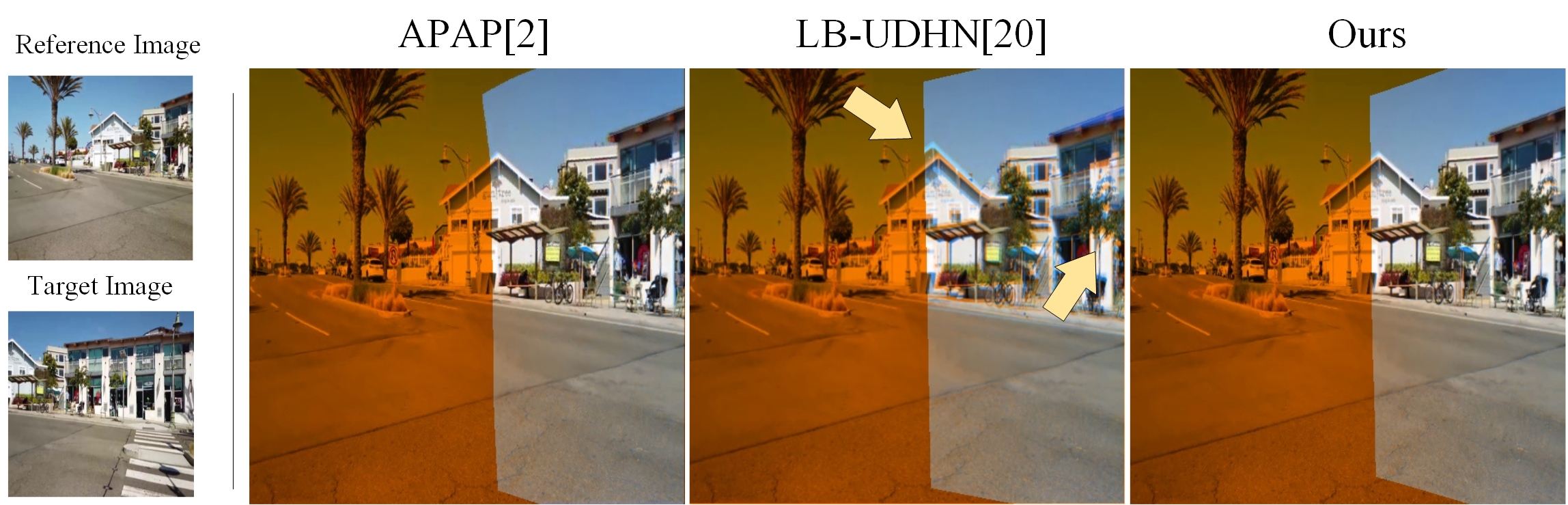}}
    \subfigure[A scene of low-texture.]
    {\includegraphics[width=0.48\textwidth]{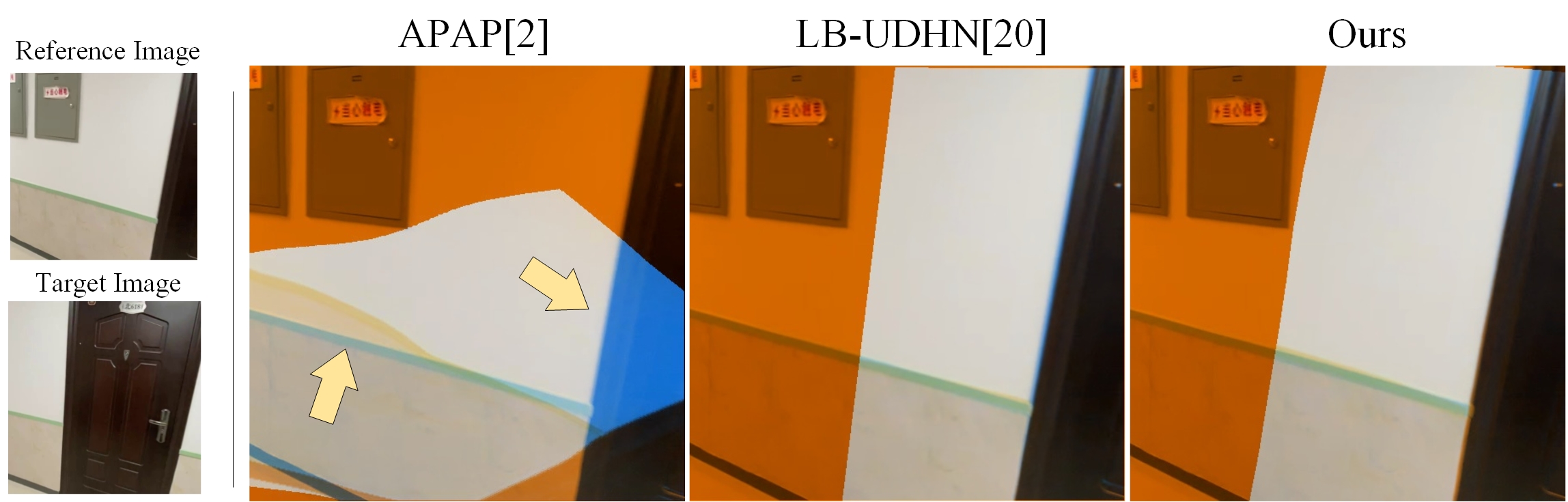}}
    \vspace{-0.2cm}
    \caption{Limitations of the existing solutions. The target image is warped to align with the reference image using the estimated homography. We fuse the reference image and the warped target image by setting the intensity of the blue channel in the reference image and that of the red channel in the warped target image to zero. \textbf{In this manner, the non-overlapping regions are distinguished in orange, which also implies the overlap rate. Overlapping regions (non-orange regions) are where we should focus, and the misalignments are highlighted in different colors.} The arrow points to the poorly aligned areas. The proposed method overcomes the two challenging cases by introducing the CCL --- a more efficient feature matching module than the cost volume.}
    %\label{fig:long}
    %\label{fig:onerow}
    \label{figure1}
 \end{figure}

Contrast to traditional algorithms, learning-based solutions \cite{detone2016deep, nguyen2018unsupervised, zhang2019content,nie2020learning, 9472883} are proposed to handle the challenging low-texture scenes for their robust deep feature extraction.
However, it is not efficient to explore the matching relationship between the deep feature maps by simply stacking convolutional layers.
Therefore, deep learning solutions often demonstrate unsatisfying performance in scenes with low overlap rates. As shown in Fig. \ref{figure1} (a), noticeable misalignments can be found in the result of the existing deep homography solution \cite{9472883} when the overlap rate is  low.

In addition, with a single estimated homography, neither traditional solutions nor deep learning ones can perfectly align images with parallax caused by different depth planes and different baselines. To address this limitation, non-linear multi-homography estimation algorithms \cite{zaragoza2013projective, lin2015adaptive, lee2020warping, 9352772} are proposed to handle the challenging parallax. Usually, these methods partition an image into dense grids, and a distinct homography can be estimated for each grid using sparse feature correspondences. However, these multi-grid methods have higher requirements on the quantity and distribution of feature correspondences. They tend to show worse performance than single homography models in low-texture or low-resolution scenes. As shown in Fig. \ref{figure1} (b), the existing traditional solution \cite{zaragoza2013projective} fails in a low-texture indoor case.

To simultaneously overcome the problems of poor robustness for traditional algorithms in low-texture scenes and failures for deep learning algorithms in low overlap rate cases, we design a contextual correlation layer (CCL) to explicitly capture the matching relationship between deep feature maps. The proposed layer is a flexible module that can be easily embedded in other matching-related networks, making a deep learning framework robust in both low-texture and low overlap rate scenarios. Compared with the cost volume \cite{hosni2012fast, sun2018pwc}, the proposed CCL has advantages of higher accuracy, faster speed, and lower memory consumption in the field of deep homography. Besides, to break the limitation that a single homography can not align images with parallax, we propose a multi-grid deep homography network to estimate multi-grid homography from global to local in an end-to-end framework. Specifically, the proposed network predicts a global homography first to align images coarsely and obtain the original shape of a mesh. Then the residual mesh deformation is estimated to enable our network to align images with parallax.

The proposed framework can be easily trained in an unsupervised manner using a pixel-level content alignment loss. However, a content constraint alone tends to cause unnatural deformations to the mesh, such as self-intersection. Existing methods \cite{wang2018deep, 10.1145/1576246.1531350} handle this problem by adding a shape-preserved constraint, encouraging all grids in the mesh to be rectangular and adjacent grids to have similar shapes. Nevertheless, this measure is essentially a trade-off between content alignment and the natural mesh shape.
In this paper, we rethink this trade-off from the perspective of parallax. A depth-aware shape-preserved loss is proposed to promote the content alignment and the natural mesh shape simultaneously.
Formally, we assign different depth levels to all grids, and only grids at the same depth level will be enforced to keep the shape consistent. As for the grids at different depth levels, there will be no shape constraint, encouraging the content alignment as much as possible. %preventing the contradiction between the content alignment and the grid shape.

Extensive experiments demonstrate that our algorithm surpasses the state-of-the-art homography estimation methods in both the synthetic benchmark dataset and the real-world dataset. Moreover, a comprehensive comparison between the proposed CCL and the cost volume is provided to support our superiority.

To sum up, we conclude our contributions as follows:
\begin{itemize}
    \item We design a novel contextual correlation layer (CCL) to explicitly explore the long-range feature correlation. It outperforms the traditional cost volume in the accuracy, the number of parameters, and the speed in the field of deep homography.
    \item We propose an end-to-end multi-grid homography estimation network, enabling our network to align images with parallax.
    \item To avoid the trade-off between the alignment and the shape, a depth-aware shape-preserved loss is presented to improve the content alignment and the natural mesh shape simultaneously.

 \end{itemize}

\section{Related Work}
\label{section2}
\subsection{Linear Homography Estimation}
\label{section21}

Pixel-based solutions iteratively estimate an optimal homography to minimize the alignment error, such as L1 norm, L2 norm, and negative of normalized correlation \cite{10.5555/1623264.1623280}. But it is challenging to handle scenarios with low overlap rates.

Different from pixel-based solutions, feature-based approaches estimate a homography by minimizing the reprojection error based on the sparse feature correspondences. The first step of these approaches is to detect sparse feature points using feature point extractors, such as SIFT\cite{lowe2004distinctive}, ORB\cite{rublee2011orb}, BRISK\cite{leutenegger2011brisk}, and SOSNet\cite{sosnet2019cvpr}. After feature matching, a robust estimation algorithm with outliers rejection will be adopted to solve for the homography from feature correspondences, such as RANSAC\cite{fischler1981random}, MAGSAC\cite{barath2019magsac}, and MAGSAC++\cite{barath2019magsacplusplus}.
Feature-based solutions work well in scenes with low overlap rates. Still, their performance heavily depends on the quality of feature correspondences, often failing in low-texture, low-light, or low-resolution scenes.

Compared with feature-based approaches, learning-based solutions address the challenge of low-texture scenes for their robust feature extraction capability. DeTone $et\ al.$ \cite{detone2016deep} apply deep learning to homography estimation for the first time, developing a VGG-style homography regression network. In their seminal work, the network predicts the 8 motions of the 4 vertices instead of solving the homography directly. By applying Spatial Transformer Network (STN) \cite{jaderberg2015spatial} to the homography network, Nguyen $et\ al.$ \cite{nguyen2018unsupervised} propose to train the regression network using a pixel-level photometric loss in an unsupervised manner. Zhang $et\ al.$ \cite{zhang2019content} and Le $et\ al.$ \cite{le2020deep} propose to reject the foreground and dynamic objects by learning a content-aware attention mask. Chang et al.\cite{8099885} design a cascaded Lucas-Kanade network to progressively refine the estimated homography in multi-scale deep features. To solve the problem of inefficient feature matching by convolution, Nie $et\ al.$ \cite{nie2020learning} connect the feature pyramid and the cost volume to regress the residual homography. Although the cost volume has significantly improved the network’s receptive field, the memory and time consumption have increased significantly.

In this paper, the proposed CCL can serve as a better alternative to the cost volume in the deep homography estimation with improved accuracy and reduced computational complexity.

\begin{figure*}[!t]
    \centering
    \includegraphics[width=1\textwidth]{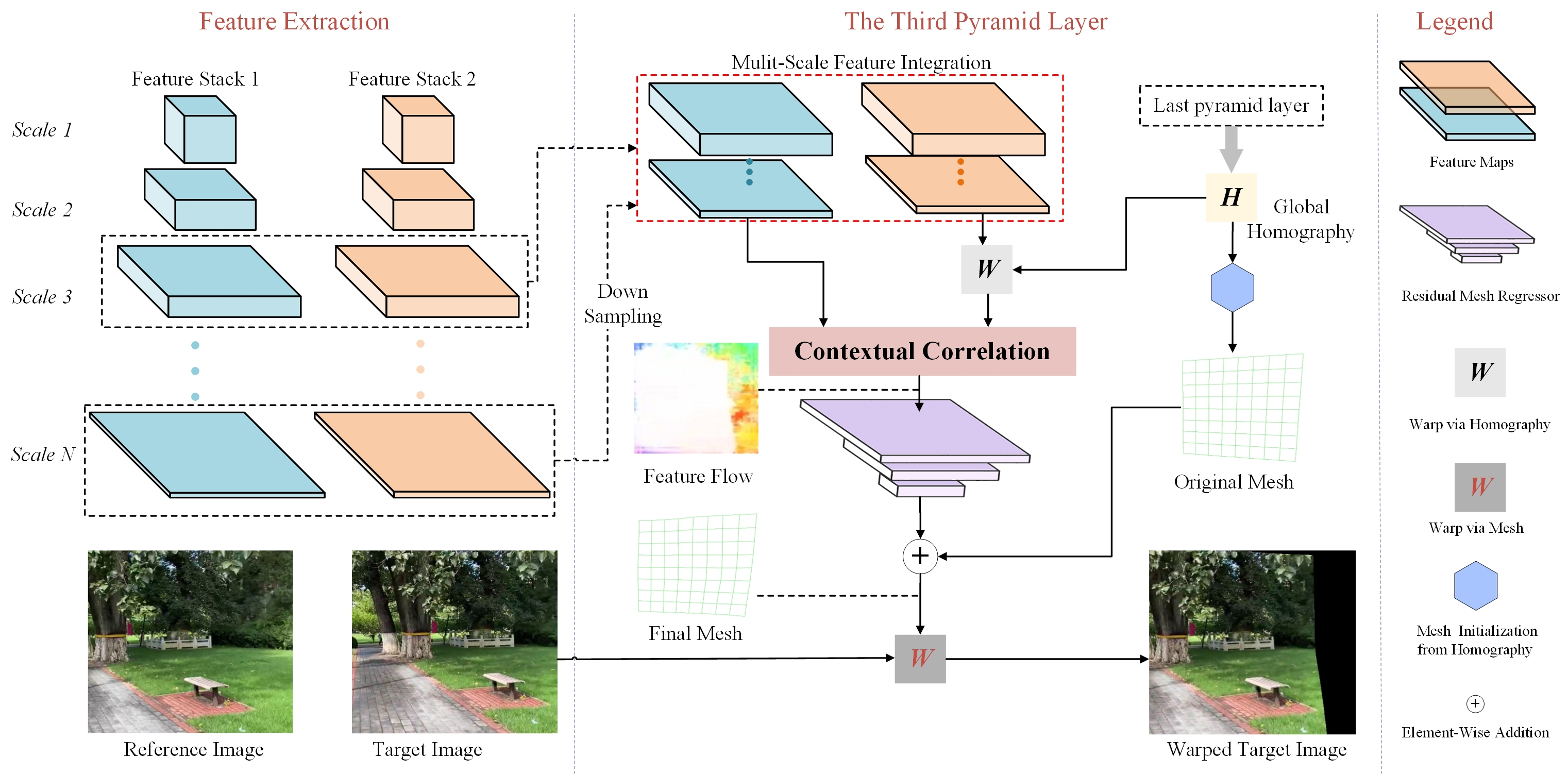}
    \vspace{-0.6cm}
    \caption{An overview of the proposed multi-grid deep homography network. The estimation can be done in a three-layer pyramid, where the first two layers predict the global homography and the third one predicts local multi-grid homography. The architecture of the third layer is demonstrated in the middle of the figure, where the multi-grid homography can be represented as the mesh.}
    \label{network}
\end{figure*}

\subsection{Non-Linear Multi-Homography Estimation}
\label{section22}
The traditional multi-homography estimations pursue precise local alignment in the overlapping regions. Zaragoza $et\ al.$ \cite{zaragoza2013projective} propose to place a mesh on the image domain and compute a local homography for every grid to achieve as-projective-as-possible (APAP) warp. To get more natural-looking results with less projective distortions, Lin $et\ al.$ \cite{lin2015adaptive} refine the local warp by connecting the homography model with similarity transformation. To get better alignment performance, Li $et\ al.$ \cite{li2017parallax} propose a robust elastic warping (robust ELA) that can be regarded as a combination of the grid-based model and the direct deformation strategy. In \cite{liao2019single}, Liao and Li propose two single-perspective warps to preserve perspective consistency with reduced projective distortions. Lee $et\ al.$ \cite{lee2020warping} partition an image into superpixels and warp them adaptively according to an optimal homography computed from the warping residual vectors. Also, Jia $et\ al.$ \cite{jia2021leveraging} leverage the line-point consistency constraint to preserve structures for wide parallax image stitching.

These traditional multi-homography solutions can achieve better alignment results than single homography solutions, especially in scenes with parallax. However, to calculate distinct local warps accurately, they have stricter requirements on the quality of feature points, often performing worse than single homography solutions in low-texture or low-resolution scenarios.

The existing learning-based multi-homography solutions are related to video stabilization, which can be seen as a small-baseline alignment application. Wang $et\ al.$ \cite{wang2018deep} propose to learn a multi-gird transformation network to align adjacent video frames. To train this network stably, the inter-grid consistency term and intra-grid regularity term are adopted to maintain the shape of the grids. In \cite{ye2019deepmeshflow}, a deep learning alternative of MeshFlow \cite{liu2016meshflow} is presented, which is content-aware and can reject dynamic objects in small-baseline scenes.
These methods eliminate the dependence on the feature points and demonstrate significantly improved robustness in low-texture scenes.

However, in applications of low overlap rates such as image stitching, these learning frameworks are not robust or even can not work. In contrast, the proposed network can solve this problem well.

\section{Methodology}
\label{section3}
\subsection{Network Overview}
\label{section31}
The proposed algorithm is a learning-based multi-grid deep homography solution, whose architecture is shown in Fig.\ref{network}. Given a reference image $I_r$ and a target image $I_t$, we aim to regress the multi-grid homography that can warp the target image to align with the reference image.
Following the previous works\cite{detone2016deep,nguyen2018unsupervised}, we represent the homography as eight motions of the four vertices which can be formulated as a matrix of size $2\times 2\times 2$. Then, the $U\times V$ multi-grid homography can be represented as a matrix of size $(U+1)\times (V+1)\times 2$.

We adopt a series of convolution-pooling blocks with shared weights to extract deep features ($F_r^k, F_t^k, k=1,2,..., N$) of the input images at different scales.
Then we resize and stack these features of different scales to formulate a feature pyramid, where the $l$-th layer integrates the multi-scale features from scale $l$ to scale $N$. Next, we use the top part of the feature pyramid ($l=1,2,3$) to predict the multi-grid homography from the integrated multi-scale features.
In our design, the $l$-th layer takes the predicted homography from the $(l-1)$-th layer as the known prior information of the current layer and predicts the residual homography. Specifically, the first two pyramid layers predict the global homography while the third pyramid layer predicts the $U\times V$ multi-grid homography.
In this way, we predict the multi-grid homography from global to local, enabling our network to process images with parallax.

\subsection{Contextual Correlation}
\label{section32}

The existing deep homography solutions do not work well in scenes with low overlap rates because the convolutional layers can not explore the long-range correlation efficiently. To solve this problem, Nie $et\ al.$\cite{nie2020learning} introduce the cost volume into the network to explicitly enhance the feature matching capability. However, the application of the cost volume significantly increases the cost of space and time. To alleviate this problem, we propose the CCL, whose structure is shown in Fig. \ref{CCL}.
It takes the feature maps of $I_r$, $I_t$ as input and outputs a dense feature flow. We define the feature flow as the motions of feature correspondences that contain the vertical and horizontal movements.

Compared with the cost volume, the proposed module has the following advantages: (1) better performance in deep homography, (2) faster speed, (3) less memory consumption,
(4) more robust on correlation matching thanks to patch-to-patch matching instead of point-to-point matching.
The first three advantages will be proved in the experiments of Section \ref{section44}, and the last one will be explained in the first step of the following implementation.

Next, we introduce three steps of the implementation:

\begin{figure}[!t]
    \centering
    \includegraphics[width=0.47\textwidth]{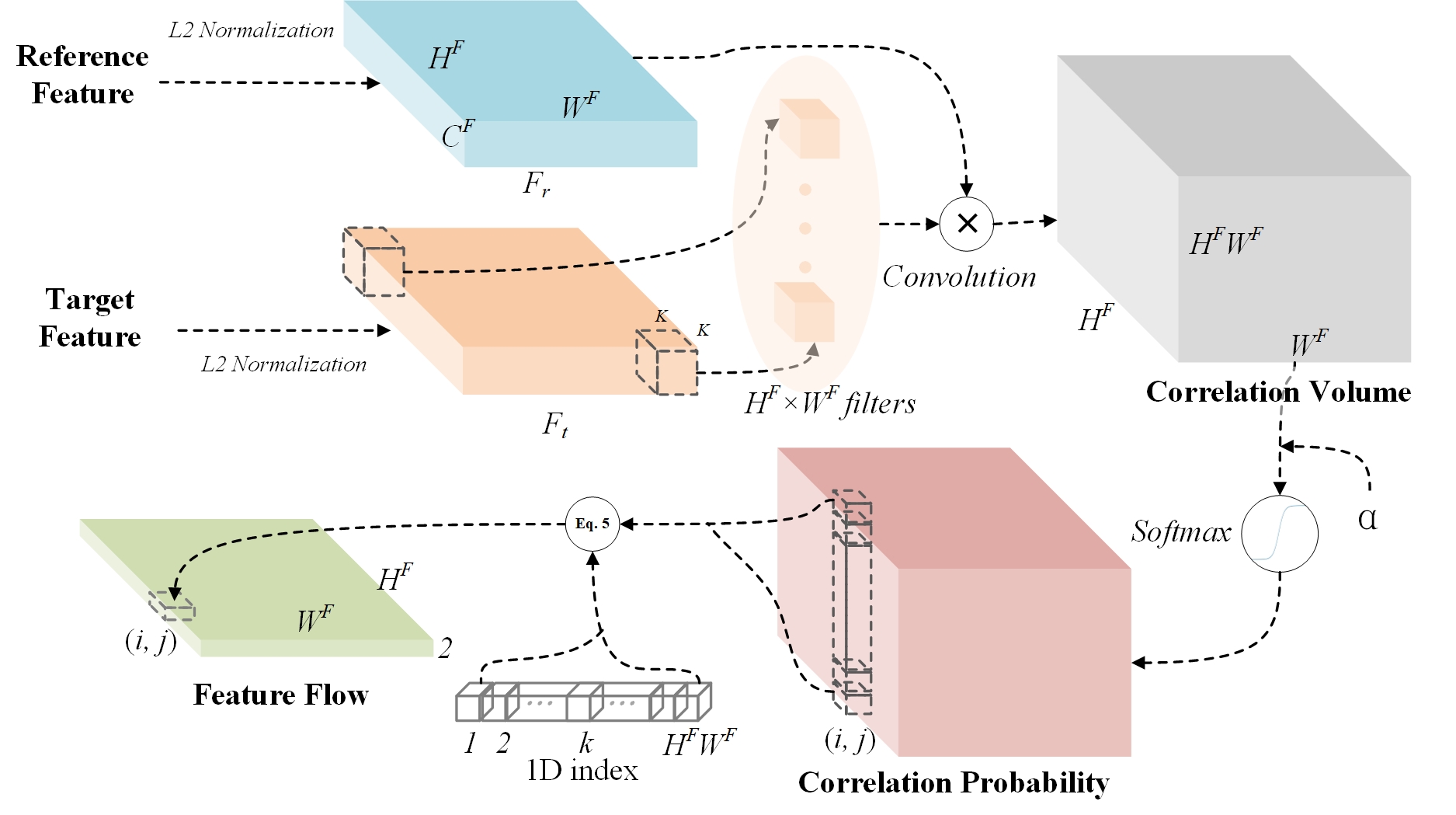}
    \vspace{-0.4cm}
    \caption{The structure of the proposed CCL. Step 1: Calculate the correlation volume by the convolution operation. Step 2: Convert the correlation volume into correlation probability. Step 3: Compute the final feature flow through Eq. \eqref{eq5}.}
    \label{CCL}
\end{figure}

\noindent\textbf{1) Correlation Volume.}

We first normalize the multi-scale features (shown in the top of Fig. \ref{network}) using $l_2$ norm on the channel dimension as $F_r, F_t\in H^F\times W^F\times C^F$ and then extract the potential correlation relationship between them.

The cost volume\cite{sun2018pwc} extracts the global correlation relationship from $F_r,F_t$ as a 3D volume with the shape of $H^F\times W^F\times (2H^F+1)(2W^F+1)$, and represents the correlation in cosine similarity as follows:
    %\vspace{-0.4cm}
\begin{equation}
    c_{x_r,y_r,x_t,y_t}=\frac{<F_r^{x_r,y_r}, F_t^{x_t,y_t}>}{|F_r^{x_r,y_r}||F_t^{x_t,y_t}|},%\ \ x_r,x_t\in \mathbb{Z}^2,
 \end{equation}
where $(x_r,y_r)$ and $(x_t,y_t)$ denote the spatial location on $F_r$ and $F_t$, respectively. $c$ is the cosine similarity between $F_r^{x_r,y_r}$ and $F_t^{x_t,y_t}$.

%但是cost volume值
Compared with the cost volume that computes the point-to-point correlation, we calculate the patch-to-patch correlation between arbitrary $K\times K$ regions in this step. As shown on the top of Fig. \ref{CCL}, we first extract dense patches ($K\times K$) from $F_t$ with the stride set to 1. Then we stack these patches as convolutional filters and use them to perform a convolutional operation on $F_r$. We call the output of the convolution the correlation volume whose shape is $H^F\times W^F\times H^FW^F$. Each value in this volume represents the similarity between a pair of arbitrary regions and can be formulated as Eq. \eqref{eq2}:

\begin{equation}
    c'_{x_r,y_r,x_t,y_t}=\sum_{i,j=-\lfloor K/2\rfloor }^{\lfloor K/2\rfloor }\hspace{-0.3cm}\frac{<F_r^{x_r+i,y_r+j}, F_t^{x_t+i,y_t+j}>}{|F_r^{x_r+i,y_r+j}||F_t^{x_t+i,y_t+j}|},%\ \ x_r,x_t\in \mathbb{Z}^2,
    \label{eq2}
 \end{equation}
where $\lfloor \cdot \rfloor $ represents the operation of round down. Compared with the shape of the cost volume, the shape of proposed correlation volume is only a quarter.

\begin{figure}[!t]
    \centering
    \includegraphics[width=0.47\textwidth]{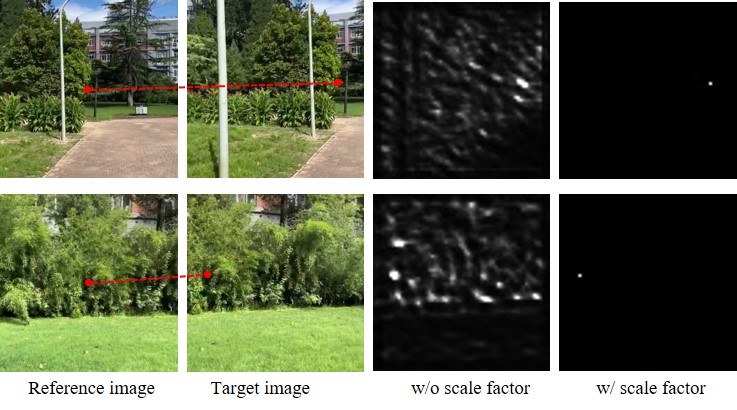}
    \vspace{-0.4cm}
    \caption{The effect of the scale factor $\alpha $. The red points in the reference/target image indicates the true matched points. In the visualization results of col 3-4, the whiter, the higher the correlation probability.}
    \label{scale_softmax}
\end{figure}

\begin{figure}[!t]
    \centering
    \subfigure[Inputs.]
    {\includegraphics[width=0.12\textwidth]{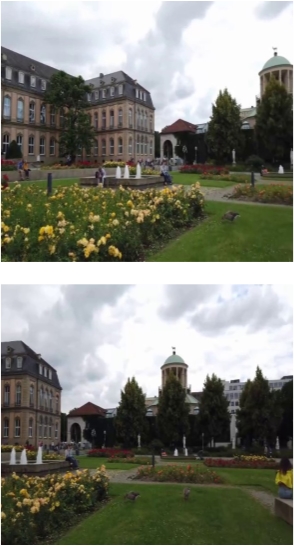}}
    \hspace{5ex}
    \subfigure[Residual feature flows.]
    {\includegraphics[width=0.27\textwidth]{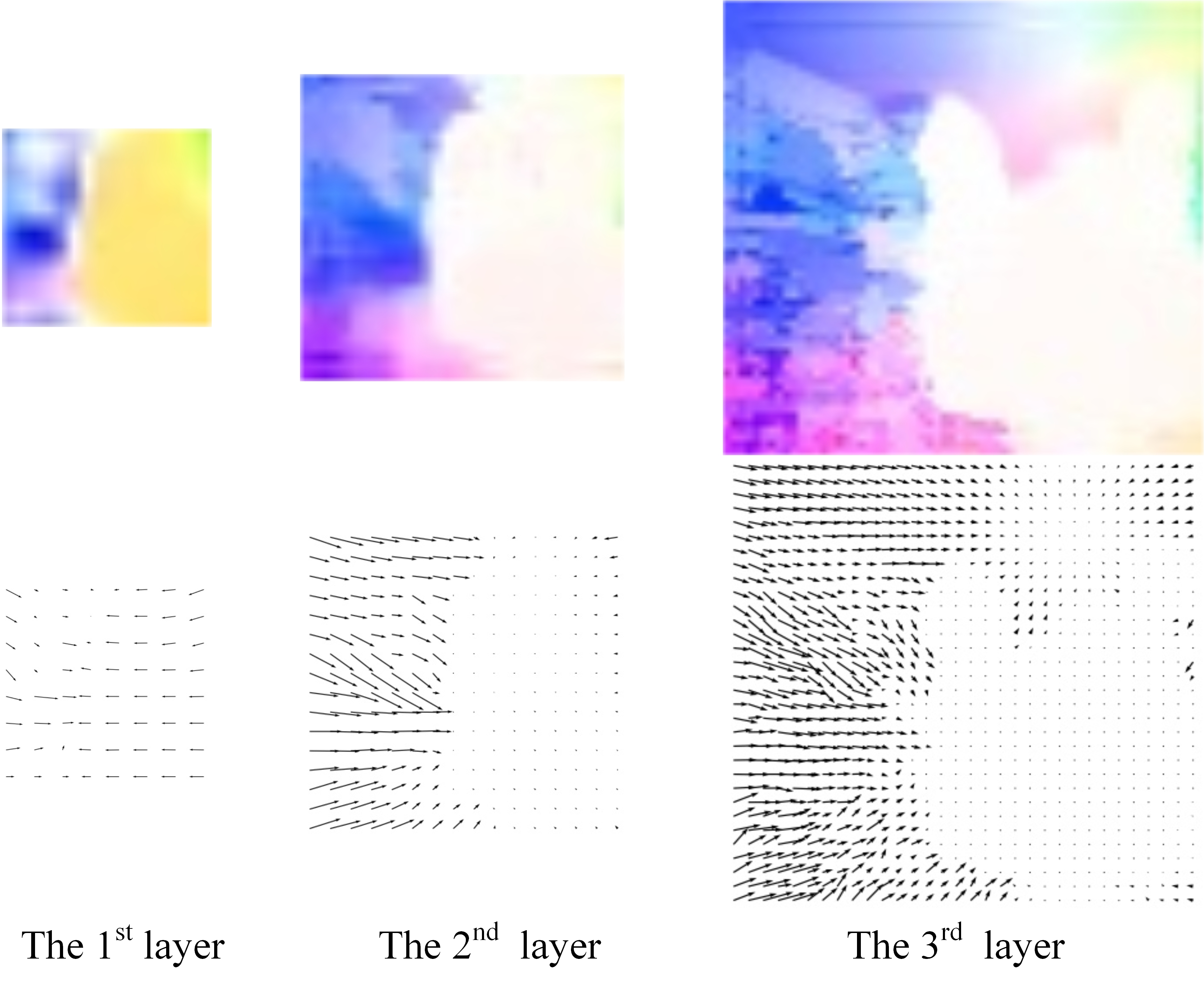}}
    \subfigure[Aligned results.]
    {\includegraphics[width=0.48\textwidth]{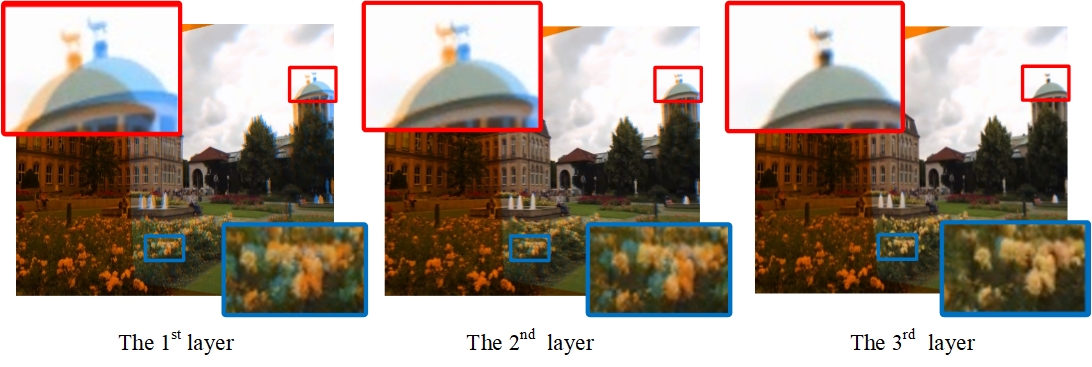}}
    \vspace{-0.2cm}
    \caption{The visualization of feature flows produced by the proposed CCL. (b) Top: the color representation. (b) Bottom: the motion representation. The network can predict the accurate homography or mesh only through the feature flows. To show the effect of the feature flows, The aligned result of every pyramid layer is also concluded.}
    %\label{fig:long}
    %\label{fig:onerow}
    \label{feature_flow}
 \end{figure}

\noindent\textbf{2) Scale Softmax.}

Every position in the correlation volume can be regarded as a vector with the length of $H^FW^F$. We set $K=3$ and every value in this vector ranges from 0 to 9. Then we use the softmax function to activate these vectors, converting the feature matching into an issue of classification ($H^FW^F$ categories). In this fashion, the correlation volume is converted to the correlation probability as illustrated in the right of Fig. \ref{CCL}. Inspired by \cite{hinton2015distilling}, these vectors will be multiplied by a constant scale factor $\alpha$ ($\alpha >1$) before the activation to increase the between-class distance.

Let the vector of correlation volume at each location  be $[x_1,x_2,...,x_{_{H^FW^F}}]^T$, and suppose $x_1\!<=\!x_2\!<=\!...\!<=\!x_{_{H^FW^F}}$. Then we can get the probability of matching of $x_k$ ($k\in \{1,2,...,H^FW^F\}$) with or without scale factor $\alpha$ as follows:

\vspace{-0.5cm}

 \begin{alignat}{2}
    p_k &\!=\!\frac{e^{x_k}}{\sum_{i=1}^{^{H^FW^F}}e^{x_i}}\!=\!\frac{1}{1\!+\!\sum_{i=1}^{k-1}e^{x_i\!-\!x_k}\!+\!\sum_{i=k+1}^{^{H^FW^F}}\!e^{x_i\!-\!x_k}},  \label{eq3}\\
    p_k^{\alpha} &\!=\!\frac{e^{\alpha x_k}}{\sum_{i=1}^{^{H^FW^F}}\!e^{\alpha x_i}}\!=\!\frac{1}{1\!+\!\sum_{i=1}^{k-1}e^{\alpha(x_i\!-\!x_k)}\!+\!\sum_{i=k+1}^{^{H^FW^F}}\!e^{\alpha(x_i\!-\!x_k)}}. \label{eq4}
   \end{alignat}

Comparing the denominator of Eq.\ref{eq3} with that of Eq.\ref{eq4}, we can observe that this scale factor can make $p_k$ with low matching probability lower (when $k$ is far less than $H^FW^F$) and $p_k$ with the highest matching probability higher (when $k=H^FW^F$ ). In other words, weak correlations are suppressed while the strongest correlation is enhanced. To further verify this view, Fig.\ref{scale_softmax} visualizes the vectors at the center of the correlation volumes after softmax with or without scale factor in 2D. It can be seen that, without the scale factor, many regions in the target image domain (col 3) are similar to the center points of the reference image. In contrast, most of the mismatched regions are rejected (col 4) with this factor.

\noindent\textbf{3) Feature Flow.}

In this step, we rethink the essence of the deep homography estimation --- regressing the 8 motions of 4 vertices in 2 orthogonal directions of the target image. Based on this comprehension, we believe that the difficulty of predicting the homography can drop if we regress the homography motions from the feature motions that indicate the dense correlations in feature maps. Hence, we propose to convert the correlation volume into feature flow --- the dense feature motions between feature correspondences at the feature level.

Let $p_k^{i,j}$ represents the value of correlation volume with scale softmax activated at position $(i,j,k)$, the feature motion$(m_{hor}^{i,j},m_{ver}^{i,j})$ at position $(i,j)$ can be calculated as follows:

\vspace{-0.2cm}
\begin{equation}
    (m_{hor}^{i,j},m_{ver}^{i,j})=\sum_{k=1}^{^{H^FW^F}}p_k^{i,j} (\rm mod\it \{k,W^F\},\lfloor k/W^F\rfloor)-(i,j),
    \label{eq5}
 \end{equation}
 where $\rm mod\{,\}$ denotes the modulus operation.

 Ultimately, we obtain the feature flow $(H^F\times W^F \times 2)$ from the reference feature to the target feature as the input of the subsequent regression network. We visualize the feature flows in Fig. \ref{feature_flow}, where the intermediate feature flows and the final aligned results in every pyramid layer are displayed in pairs.

 \begin{table}[]
    \centering
    \caption{Comparison of complexity between the cost volume and proposed contextual correlation. Suppose $H^F=W^F=n$.}
\scalebox{0.85}{
    \begin{tabular}{|l||l|l|}

        \hline
         & \makecell[c]{Cost volume}& \makecell[c]{Contextual correlation}\\
        \hline
        \hline
        \makecell[c]{Implementation}& \makecell[c]{Loop} & \makecell[c]{Convolution} \\
        \hline
        \makecell[c]{Space complexity}& \makecell[c]{$O(n^4)$}& \makecell[c]{$O(n^4)$}  \\
        \hline
        \makecell[c]{Time complexity}& \makecell[c]{$O(n^2)$} & \makecell[c]{$O(1)$} \\
        \hline

  \end{tabular}
}
\label{complexity}
\end{table}

Also, a comparison of the complexity between the cost volume and our CCL is provided in Table \ref{complexity}.
Compared with the cost volume, the CCL outputs the lightweight representation of feature flows by rejecting most unmatched positions, leading to more efficient prediction.

 \begin{figure}[!t]
    \centering
    \subfigure[Input images.]
    {\includegraphics[width=0.18\textwidth]{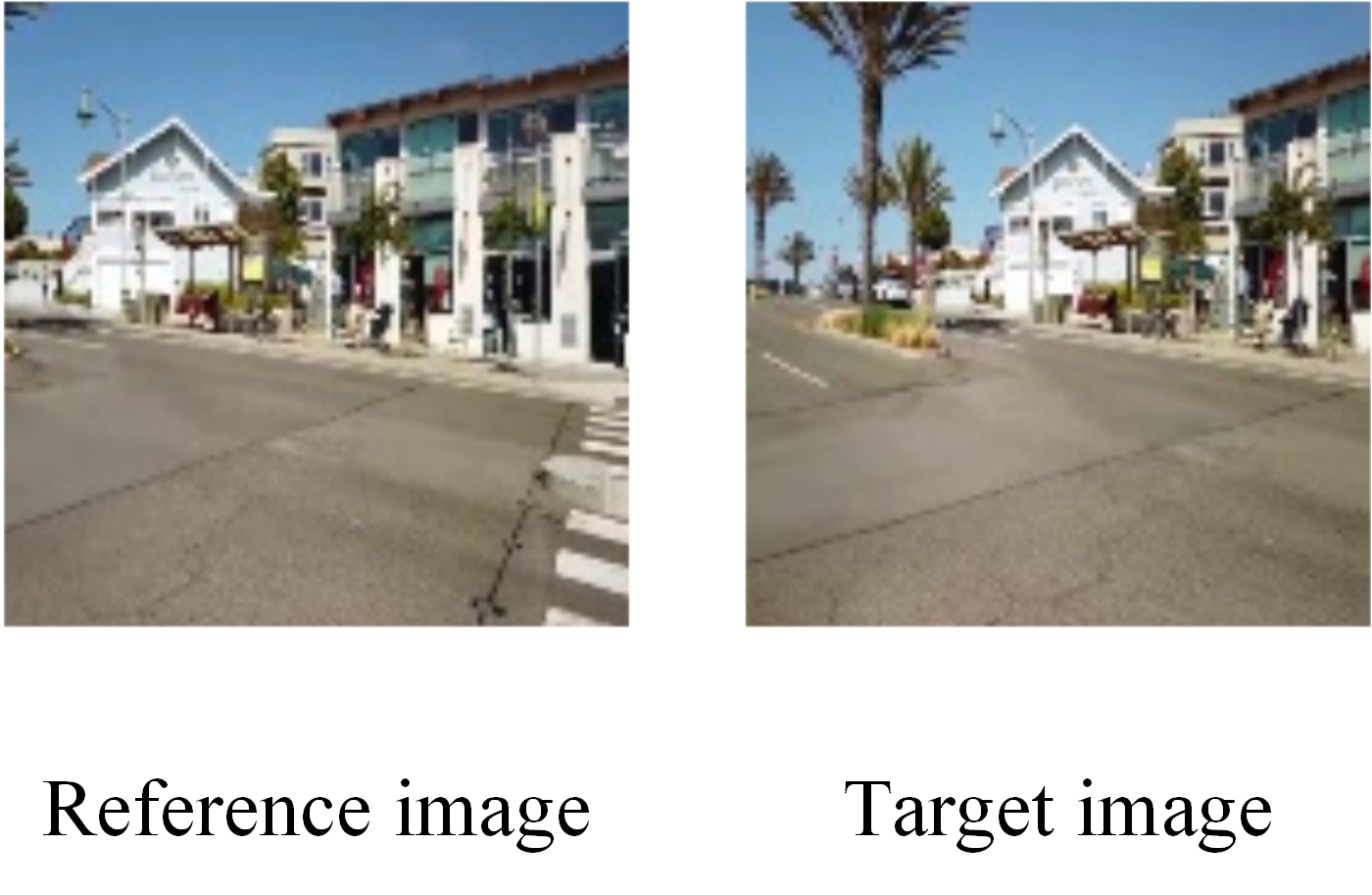}}
    \subfigure[Forward single grid deformation.]
    {\includegraphics[width=0.28\textwidth]{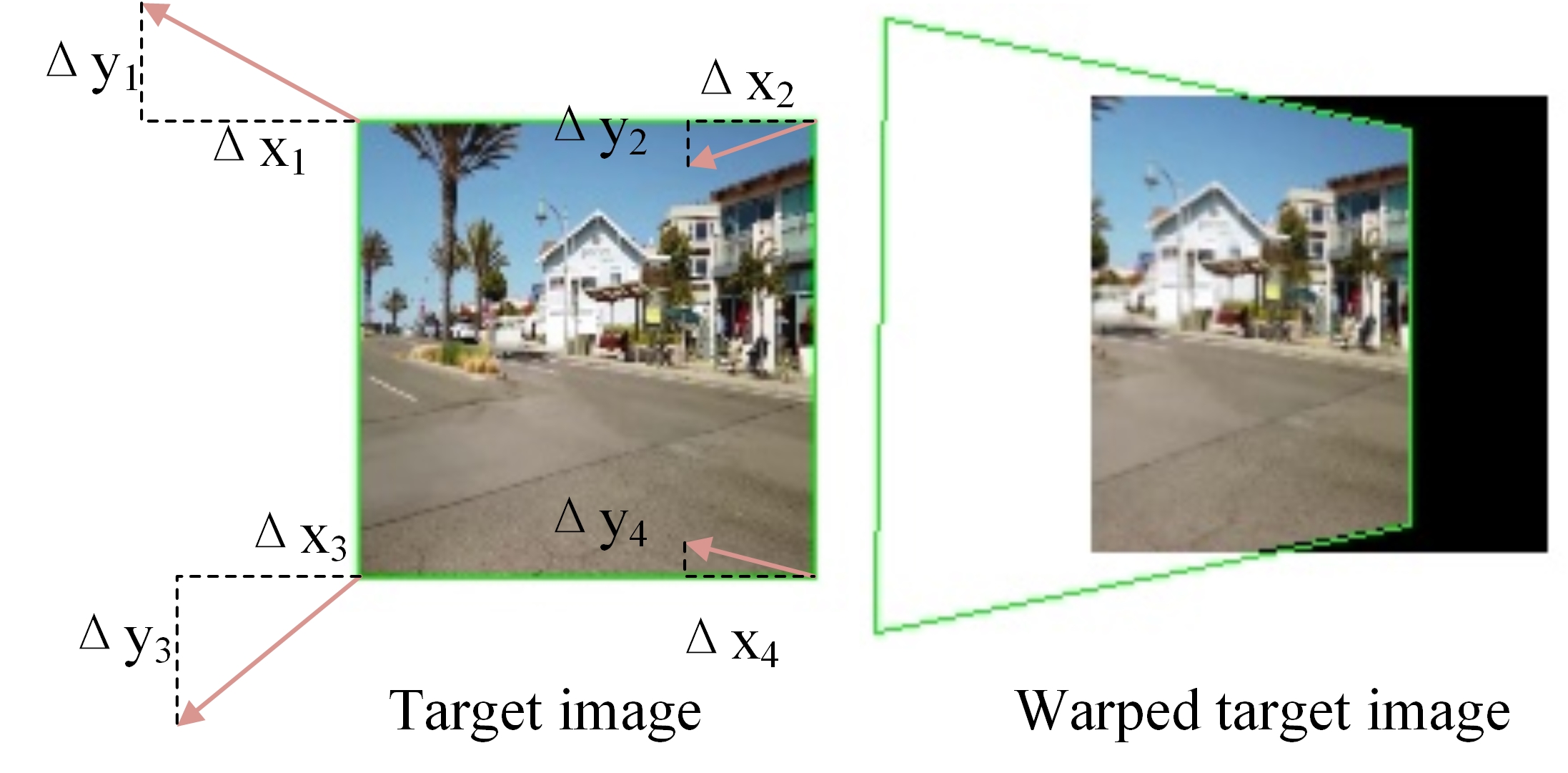}}
    \subfigure[Forward multi-grid deformation.]
    {\includegraphics[width=0.23\textwidth]{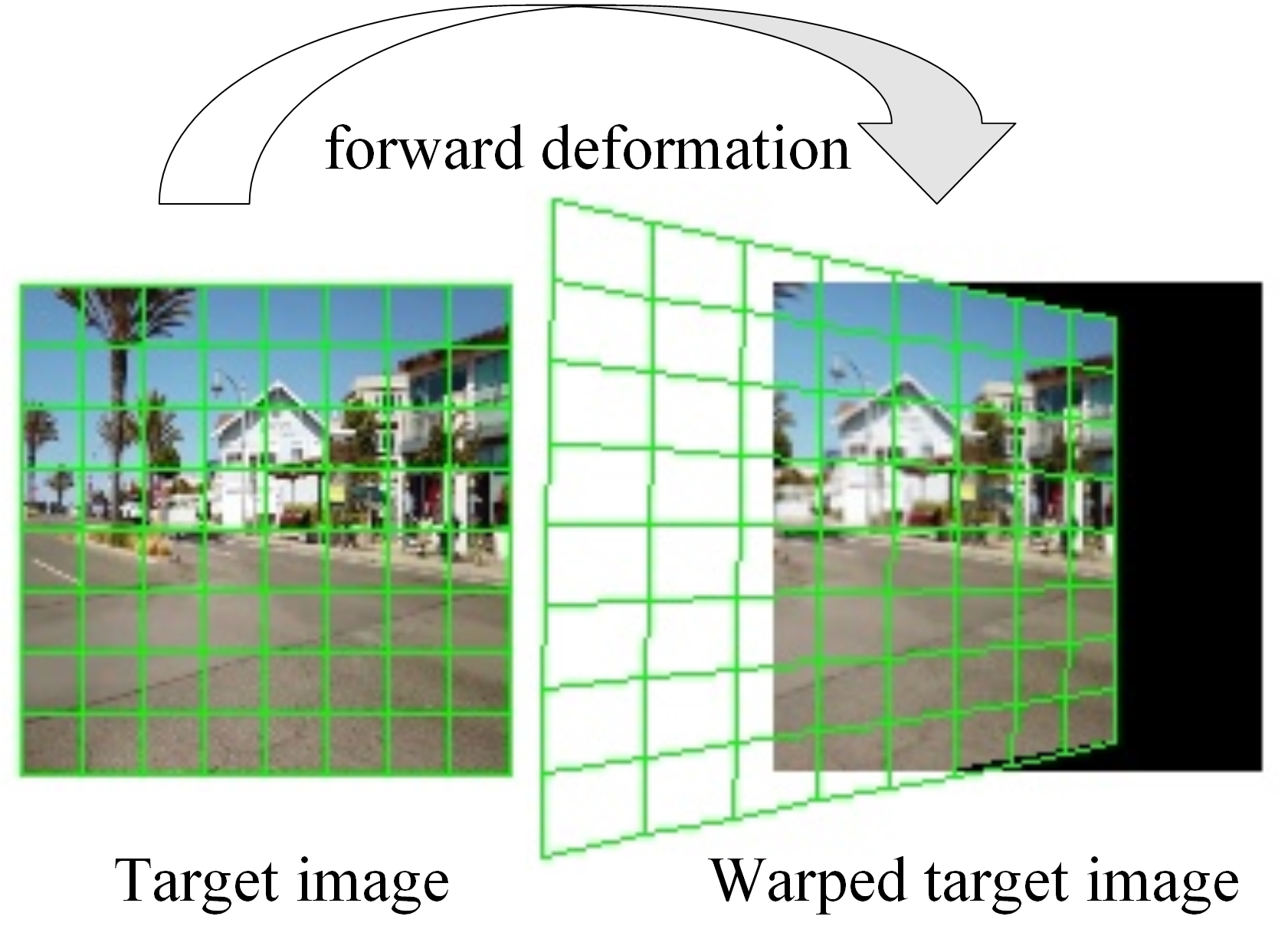}}
    \subfigure[Backward multi-grid deformation.]
    {\includegraphics[width=0.24\textwidth]{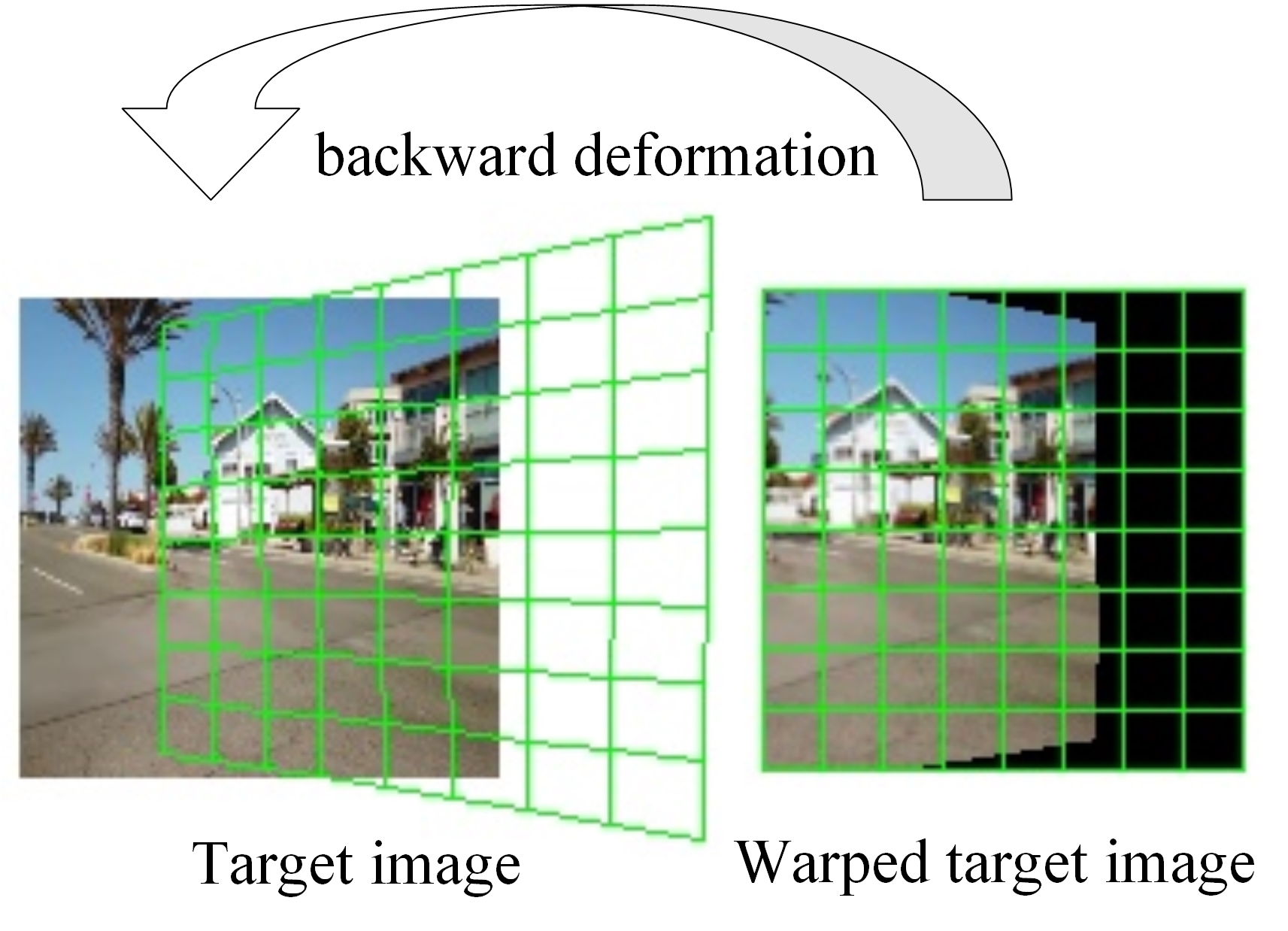}}
    \vspace{-0.2cm}
    \caption{Comparison between the forward deformation and backward deformation.}
    %\label{fig:long}
    %\label{fig:onerow}
    \label{backward_deformation}
 \end{figure}

 \subsection{Backward Multi-Grid Deformation}
 \label{section33}

The existing deep homography solutions predict the 8 motions of the 4 vertices in the target image instead of directly solving the 8 unknown parameters in the homography. With the 8 motions, we can solve for the corresponding homography and warp the target image to align with the reference image as shown in Fig. \ref{backward_deformation} (a)(b). We call this motion direction from the target image domain to the reference image domain as the forward deformation. Then, by placing a mesh on the target image, we can extend the single grid deformation ($1\times 1$) to multi-grid deformation ($U\times V$) as shown in Fig. \ref{backward_deformation} (c). Different from single grid warping in which every pixel shares the same homography, multi-grid warping has to assign a different homography to each pixel in the warped target image. In other words, in multi-grid warping, we need to figure out which grid the pixel in the warped target image belongs to. Since the mesh shape in the warped target image is irregular after the forward deformation, it is hard to obtain the correspondence between each grid and each pixel in the warped target image using a fast and efficient way in a deep learning implementation. Therefore, if we continue to adopt the forward deformation, it can drastically reduce the speed.

To avoid this problem, we design a backward deformation solution. Specifically, as shown in Fig. \ref{backward_deformation} (d), we place a regular mesh in the warped target image and then predict the grid motions from the reference image domain to the target image domain.
Compared with the forward deformation, the mesh shape in the warped target image is regular in the proposed backward deformation. Therefore, we can easily assign different homography to each pixel of the warped target image in batches, enabling parallel acceleration in GPU.

 \begin{figure}[!t]
    \centering
    \subfigure[Depth perception on the warped target image.]
    {\includegraphics[width=0.48\textwidth]{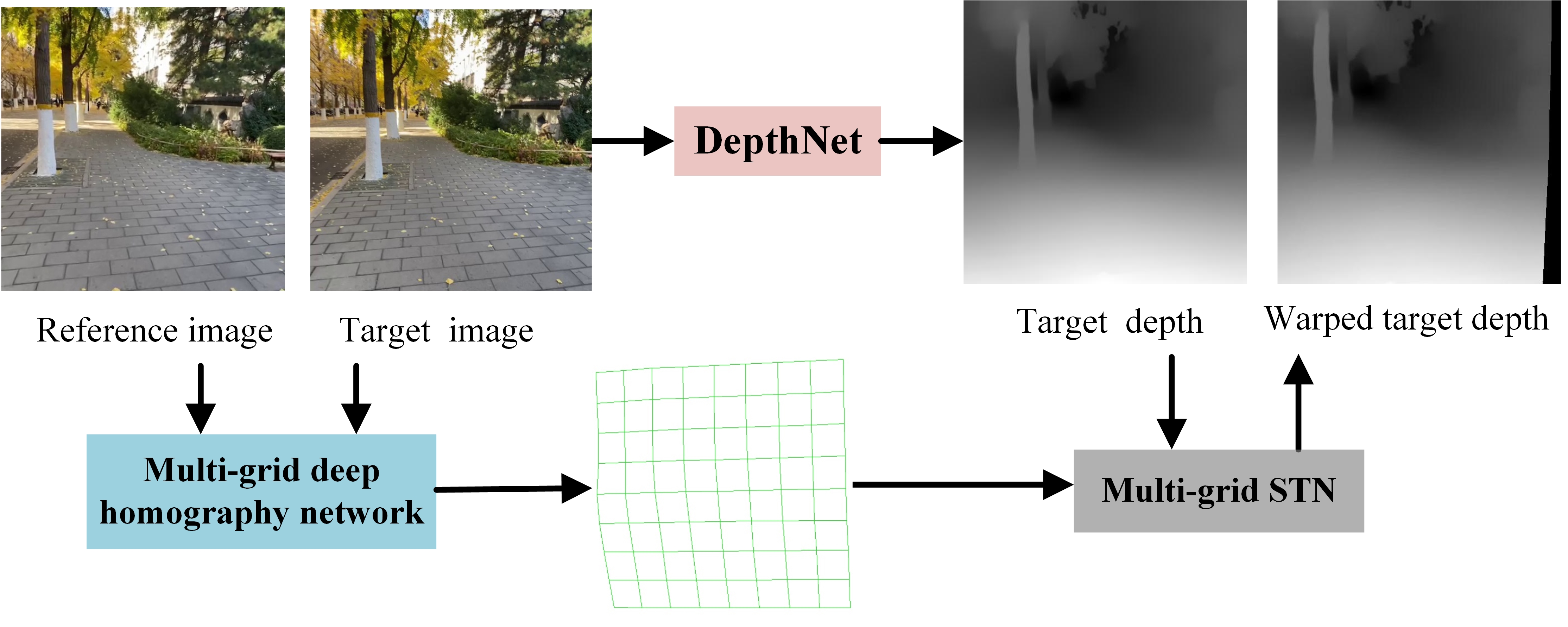}}
    \subfigure[Depth levels Visualization.]
    {\includegraphics[width=0.31\textwidth]{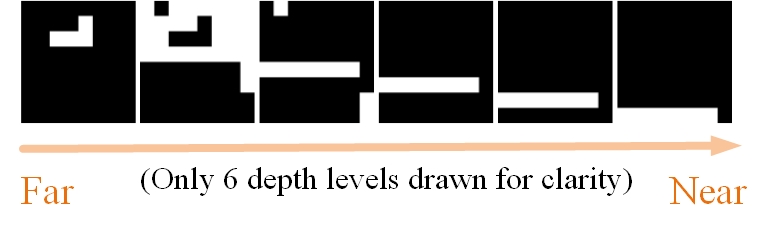}}
    \subfigure[Shape-preserved grids.]
    {\includegraphics[width=0.165\textwidth]{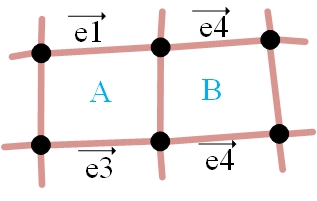}}
    \vspace{-0.2cm}
    \caption{Depth-aware shape-preserved loss. (a) Depth perception from a pretrained monocular depth estimation network (b) Depth levels division on the mesh. (c) The shape-preserved constraint on adjacent edges in the same depth level.}
    %\label{fig:long}
    %\label{fig:onerow}
    \label{depth_aware}
 \end{figure}

 \subsection{Unsupervised Training}
 \label{section34}
The proposed multi-grid deep homography network is trained in a fully unsupervised manner since there is no ground truth in real-world scenes. In this paper, we design a content alignment term and a shape-preserved term to optimize our network. The content alignment term is adopted to align the input images in image contents, while the shape-preserved term is designed to keep the mesh shape from unnatural distortions.

\begin{spacing}{1.5}
\end{spacing}
\textbf{1) Content Alignment Loss.}

 %\subsubsection{Content Alignment Loss}
 %\label{section341}
 Given a reference image $I_r$ and a target image $I_t$ with overlapping regions, the network is expected to align the overlapping regions. Let $\mathcal{W}^k(\cdot)$ be the operation of spatial transformation using the predicted mesh in the $k$-th layer of the feature pyramid. Then the content alignment constraint in this layer can be written as follows:
 \begin{equation}
    L_{content}^k=\|\mathcal{W}^k(E)\odot I_r - \mathcal{W}^k(I_t)\Vert_1 ,
 \end{equation}
where $E$ is an all-one matrix and $\odot$ is the operation of element-wise multiplication.

Considering our network consists of 3 pyramid layers, the content alignment loss should be:
\begin{equation}
    L_{content}=\omega_1L_{content}^1+\omega_2L_{content}^2+\omega_3L_{content}^3,
 \end{equation}
 where $\omega_1$, $\omega_2$, and $\omega_3$ are the weights for different pyramid layers.

 \begin{spacing}{1.5}
 \end{spacing}
 \textbf{2) Depth-Aware Shape-Preserved Loss.}
 %\subsubsection{Depth-Aware Shape-Preserved Loss}
 %\label{section341}

Optimizing the network using content alignment loss alone can cause unnatural mesh distortions, such as self-intersection. To avoid this problem, a shape-preserved loss is proposed in \cite{ye2019deepmeshflow,wang2018deep}, which encourages adjacent grids to maintain similar shapes. However, this constraint can easily spread from adjacent grids to the surroundings, enforcing all the grids to maintain a similar shape. This manner treats all grids as being on the same plane, improving the shape regularity at the cost of reducing the content alignment performance.

In this paper, we rethink this shape-preserved loss: the ideal loss should only impose the shape-preserved constraint on grids at the same depth, but not on grids at different depths. To implement it, we design a depth-aware shape-preserved loss, which can estimate different depth levels on the mesh and constrain the grids at the same depth level to maintain similar shapes.

The process of calculating the depth-aware shape-preserved loss is shown in Fig. \ref{depth_aware}. First, we adopt a pre-trained monocular depth estimation model \cite{Xian_2020_CVPR} to predict the depth map of the target image. Then, we warp the depth map using the multi-grid homography estimated from our network. Having obtained the warped depth maps, we compute the average depth value in every grid of the warped depth map. Next, we divide the warped depth map into $M$ different levels ($D^k, k=1,2,..., M$) with the same intervals.
 %with every depth level ranging form $\frac{1}{M}i$ to $\frac{1}{M}(1+i)$ ($i=0,1,...,M-1$).
The depth levels are visualized in Fig. \ref{depth_aware} (b), where the white indicates these regions are at the same depth.

Assuming all the grids are at the same depth level, we can warp them with the same homography, and each line in the deformated mesh is a straight line. Based on this observation, we constrain the direction of grid edge consistent in adjacent grids at the same depth. An example is shown in Fig. \ref{depth_aware} (c), where the similarity between grid A and B can be formulated as follows:
 \begin{equation}
    l_{sp}^{A,B}= 2-\frac{\left\lvert \overrightarrow{e_1}\cdot \overrightarrow{e_2}\right\rvert }{\left\lVert \overrightarrow{e_1}\right\rVert \cdot \left\lVert \overrightarrow{e_2}\right\rVert } - \frac{\left\lvert \overrightarrow{e_3}\cdot \overrightarrow{e_4}\right\rvert }{\left\lVert \overrightarrow{e_3}\right\rVert \cdot \left\lVert \overrightarrow{e_4}\right\rVert }.
    \label{sp}
 \end{equation}
According to Eq. \eqref{sp}, we can calculate the similarity matrices ($L_{sp}^{hor}$ and $L_{sp}^{ver}$) on horizontal and vertical directions of the deformated mesh, where the sizes of $L_{sp}^{hor}$ and $L_{sp}^{ver}$ are $U\times (V-1)$ and $(U-1)\times V$, respectively. Finally, we formulate our depth-aware shape-preserved loss as follows:
 \begin{equation}
    L_{shape}=\frac{1}{U(V-1)}\sum_{k=1}^{M} D_{hor}^kL_{sp}^{hor} + \frac{1}{(U-1)V}\sum_{k=1}^{M} D_{ver}^kL_{sp}^{ver},
 \end{equation}
 where $D_{hor}^k$ and $D_{ver}^k$ are depth consistency matrices on the horizontal and vertical directions, and they are calculated from $D^k$. Every element in $D_{hor}^k$ ($U\times (V-1)$) and $D_{ver}^k$ ($(U-1)\times V$) indicates if the adjacent grids (horizontal direction or vertical direction) are at the same depth level.

 \begin{spacing}{1.5}
 \end{spacing}
 \textbf{3) Objective Function.}

 %\subsubsection{Objective Function}
 %\label{section342}
Taking the content alignment term and shape-preserved term into consideration simultaneously, we conclude the objective function of our network:
 \begin{equation}
    L =  \lambda L_{content}+ \mu L_{shape},
 \end{equation}
 where $\lambda$ and $\mu$ are the weights of $L_{content}$ and $L_{shape}$, respectively.

 \section{Experimental Results}
\label{section4}

\subsection{Datasets and Implement Details}
\label{section41}
\textbf{Datasets.}
We validate the performance of the proposed network in two public datasets. The first one is a synthetic benchmark dataset that is called Warped MS-COCO \cite{detone2016deep}. The samples in this dataset are image pairs without parallax, thus the target image can be aligned with the reference image perfectly using a single homography. The second dataset is a real-world dataset that is proposed in an unsupervised deep image stitching work \cite{9472883}, where 10,440 image pairs are used for training and 1,106 image pairs are used for testing. This dataset, called UDIS-D, consists of varying overlap rates, different degrees of parallax, and variable scenes such as indoor, outdoor, night, dark, snow, and zooming.

\textbf{Implement Details.}
Our network is trained using an Adam \cite{kingma2014adam} optimizer with an exponentially decaying learning rate initialized to $10^{-4}$ for 500$k$ iterations. The batch size is set to 4, and we first train the network for 300$k$ iterations with $\lambda$ and $\mu$ set to 1 and 0, respectively. For the remaining 200$k$ iterations, we set $\lambda$ and $\mu$ to 1 and 10. And $\omega_1$, $\omega_2$, $\omega_3$ and $\alpha$ are assigned as 1, 4, 16, and 10. We use RELU as the activation function for all the convolutional layers except that the last layer in the regression network adopts no activation function. The implementation is based on TensorFlow and the network is performed on a single GPU with NVIDIA RTX 2080 Ti. It takes about 96$ms$ to align the images of 512$\times$512 resolution.

\subsection{Comparison on Synthetic Benchmark Dataset}
\label{section42}

\begin{table}[]
    \centering
    \caption{Comparison experiment on Warped MS-COCO \cite{detone2016deep}. The 1st and 2nd best solutions are marked in red and blue, respectively.}

\scalebox{0.75}{
    \begin{tabular}{|l||l|l|l|l|}
        \hline
        \makecell[c]{\multirow{2}{*}{}} & \multicolumn{4}{c|}{4-pt Homography RMSE ($\downarrow$)}\\
        \cline{2-5}
        & \makecell[c]{Easy} & Moderate & Hard & \makecell[c]{Average}\\
        \hline
        \hline
        \makecell[c]{$I_{3\times 3}$} & \makecell[c]{15.0154} & \makecell[c]{18.2515} & \makecell[c]{21.3548} & \makecell[c]{18.5220}\\
        \hline
        \hline
        \makecell[c]{SIFT\cite{lowe2004distinctive}+RANSAC\cite{fischler1981random}} & \makecell[c]{0.6687} & \makecell[c]{1.1223} & \makecell[c]{18.5990} & \makecell[c]{7.9769}\\
        \hline
        \makecell[c]{ORB\cite{rublee2011orb}+RANSAC\cite{fischler1981random}} & \makecell[c]{3.8995} & \makecell[c]{10.2206} & \makecell[c]{$F$} & \makecell[c]{$F$}\\
        \hline
        \makecell[c]{BRISK\cite{leutenegger2011brisk}+RANSAC\cite{fischler1981random}} & \makecell[c]{1.2179} & \makecell[c]{2.6831} & \makecell[c]{$F$} & \makecell[c]{$F$}\\
        \hline
        \makecell[c]{SOSNet\cite{sosnet2019cvpr}+RANSAC\cite{fischler1981random}} & \makecell[c]{0.6228} & \makecell[c]{1.0162} & \makecell[c]{15.7217} & \makecell[c]{6.7804}\\
        \hline
        \makecell[c]{SIFT\cite{lowe2004distinctive}+MAGSAC\cite{barath2019magsac}} & \makecell[c]{0.5697} & \makecell[c]{0.8679} & \makecell[c]{$F$} & \makecell[c]{$F$}\\
        \hline
        \makecell[c]{ORB\cite{rublee2011orb}+MAGSAC\cite{barath2019magsac}} & \makecell[c]{3.1557} & \makecell[c]{8.7443} & \makecell[c]{$F$} & \makecell[c]{$F$}\\
        \hline
        \makecell[c]{BRISK\cite{leutenegger2011brisk}+MAGSAC\cite{barath2019magsac}} & \makecell[c]{1.1472} & \makecell[c]{2.6300} & \makecell[c]{$F$} & \makecell[c]{$F$}\\
        \hline
        \makecell[c]{SOSNet\cite{sosnet2019cvpr}+MAGSAC\cite{barath2019magsac}} & \makecell[c]{0.5790} & \makecell[c]{0.9006} & \makecell[c]{$F$} & \makecell[c]{$F$}\\
        \hline
        \makecell[c]{SIFT\cite{lowe2004distinctive}+MAGSAC++\cite{barath2019magsacplusplus}} & \makecell[c]{0.5663} & \makecell[c]{0.8388} & \makecell[c]{$F$} & \makecell[c]{$F$}\\
        \hline
        \makecell[c]{ORB\cite{rublee2011orb}+MAGSAC++\cite{barath2019magsacplusplus}} & \makecell[c]{2.7802} & \makecell[c]{7.7074} & \makecell[c]{$F$} & \makecell[c]{$F$}\\
        \hline
        \makecell[c]{BRISK\cite{leutenegger2011brisk}+MAGSAC++\cite{barath2019magsacplusplus}} & \makecell[c]{1.0144} & \makecell[c]{2.2898} & \makecell[c]{$F$} & \makecell[c]{$F$}\\
        \hline
        \makecell[c]{SOSNet\cite{sosnet2019cvpr}+MAGSAC++\cite{barath2019magsacplusplus}} & \makecell[c]{0.5564} & \makecell[c]{0.8139} & \makecell[c]{$F$} & \makecell[c]{$F$}\\
        \hline
        \hline
        \makecell[c]{DHN\cite{detone2016deep}} & \makecell[c]{3.2998} & \makecell[c]{4.8839} & \makecell[c]{7.7017} & \makecell[c]{5.5358}\\
        \hline
        \makecell[c]{UDHN\cite{nguyen2018unsupervised}} & \makecell[c]{2.1894} & \makecell[c]{3.5272} & \makecell[c]{6.5073} & \makecell[c]{4.3179}\\
        \hline
        \makecell[c]{CA-DHN\cite{zhang2019content}} & \makecell[c]{15.0082} & \makecell[c]{18.2498} & \makecell[c]{$F$} & \makecell[c]{$F$}\\
        \hline
        \makecell[c]{LB-DHN\cite{nie2020learning}} & \makecell[c]{$\mathbf{\textcolor{blue}{0.2719}}$} & \makecell[c]{$\mathbf{\textcolor{blue}{0.4140}}$} & \makecell[c]{$\mathbf{\textcolor{blue}{0.9761}}$} & \makecell[c]{$\mathbf{\textcolor{blue}{0.5962}}$}\\
        \hline
        \makecell[c]{LB-UDHN\cite{9472883}} & \makecell[c]{1.1773} & \makecell[c]{1.4544} & \makecell[c]{3.0702} & \makecell[c]{2.0239}\\
        \hline
        \hline
        \makecell[c]{Ours} & \makecell[c]{$\mathbf{\textcolor{red}{0.2163}}$} & \makecell[c]{$\mathbf{\textcolor{red}{0.3349}}$} & \makecell[c]{$\mathbf{\textcolor{red}{0.7076}}$} & \makecell[c]{$\mathbf{\textcolor{red}{0.4484}}$}\\
        \hline
  \end{tabular}
}
\label{synthetic_dataset}
\end{table}

Since Warped MS-COCO is a synthetic dataset without parallax, we only compare our solution with other single homography solutions. Besides, we modify the third pyramid level of the regression network to predict a single homography instead of the multi-homography for fairness. We train our network in a supervised manner because of the availability of the ground truth.

We first compare ours with traditional feature-based solutions. As shown in Table \ref{synthetic_dataset}, we choose various feature descriptors, $e.g.$ SIFT \cite{lowe2004distinctive}, ORB \cite{rublee2011orb}, BRISK \cite{leutenegger2011brisk} and SOSNet \cite{sosnet2019cvpr}, and different outliers rejection algorithms, $e.g.$ RANSAC \cite{fischler1981random}, MAGSAC \cite{barath2019magsac}, MAGSAC++ \cite{barath2019magsacplusplus}, composing 12 distinct solutions. Of these feature descriptors, SOSNet is adopted to replace the local descriptors of SIFT with deep learning features.

Then we compare ours with deep learnig-based solutions, $e.g.$ DHN \cite{detone2016deep}, UDHN \cite{nguyen2018unsupervised}, CA-DHN \cite{zhang2019content}, LB-DHN \cite{nie2020learning} and LB-UDHN \cite{9472883}. All the learning solutions including ours are trained in this dataset.

We adopt 4-pt homography RMSE from \cite{nguyen2018unsupervised, 9472883} as the metric to evaluate all the above solutions. Formally, we divide the testing results into 3 levels according to the performance --- easy (top 0-30\%), moderate (top 30-60\%), and hard (top 60-100\%). $I_{3\times 3}$ refers to a $3\times 3$ identity matrix as a ``no-warping' homography for the reference.

The results are shown in Table \ref{synthetic_dataset}, where $F$ indicates the performance of the solution is worse than that of $I_{3\times 3}$. Observed from this table, it is evident to conclude that:

(1) Our network outperforms other solutions, including traditional and deep learning algorithms in all cases (``easy'', ``moderate'', and ``hard'').

(2) The deep learning solutions are more robust since they are superior to traditional solutions in ``hard'' columns. This advantage benefits from the robust feature extraction ability of CNNs.

In the experiment, we do not compare ours with \cite{8099885, zhao2021deep} because they are essentially template matching algorithms. Specifically, they will take a big reference image and a small template image as inputs in these methods, where the overlap rates between the inputs can be 100\%. In contrast, these compared methods and ours deal with inputs with a relatively low overlap rate.

\begin{table*}[]
    \centering
    \caption{Quantitative comparison with state-of-the-arts on real-world dataset (UDIS-D). The 1st and 2nd best solutions are marked in red and blue, respectively.}
    \subtable[Quantitative comparison on low-resolution.]{
        \scalebox{0.9}{
            \begin{tabular}{|l||l|l|l|l||l|l|l|l|}
                \hline
                \makecell[c]{\multirow{3}{*}{}}  & \multicolumn{8}{c|}{$128\times 128$}  \\
                \cline{2-9}
                & \multicolumn{4}{c||}{PSNR ($\uparrow$)} & \multicolumn{4}{c|}{SSIM ($\uparrow$)} \\
                \cline{2-9}
                & \makecell[c]{Easy} & Moderate & Hard & \makecell[c]{Average} & \makecell[c]{Easy} & Moderate & Hard & \makecell[c]{Average} \\
                \hline
                \hline
                \makecell[c]{$I_{3\times 3}$} & \makecell[c]{16.19} & \makecell[c]{13.05} & \makecell[c]{10.87} & \makecell[c]{13.12} & \makecell[c]{0.397} & \makecell[c]{0.173} & \makecell[c]{0.065} & \makecell[c]{0.197} \\
                \hline
                \hline
                \makecell[c]{SIFT\cite{lowe2004distinctive}+RANSAC\cite{fischler1981random}} & \makecell[c]{25.23} & \makecell[c]{22.23} & \makecell[c]{17.53} & \makecell[c]{21.25} & \makecell[c]{0.860} & \makecell[c]{0.766} & \makecell[c]{0.558} & \makecell[c]{0.711} \\
                \hline
                \makecell[c]{ORB\cite{rublee2011orb}+RANSAC\cite{fischler1981random}} & \makecell[c]{21.94} & \makecell[c]{18.58} & \makecell[c]{13.49} & \makecell[c]{17.55} & \makecell[c]{0.734} & \makecell[c]{0.575} & \makecell[c]{0.361} & \makecell[c]{0.537}\\
                \hline
                \makecell[c]{BRISK\cite{leutenegger2011brisk}+RANSAC\cite{fischler1981random}} & \makecell[c]{24.38} & \makecell[c]{21.53} & \makecell[c]{16.14} & \makecell[c]{20.23} & \makecell[c]{0.832} & \makecell[c]{0.735} & \makecell[c]{0.502} & \makecell[c]{0.671}\\
                \hline
                \makecell[c]{SOSNet\cite{sosnet2019cvpr}+RANSAC\cite{fischler1981random}} & \makecell[c]{25.35} & \makecell[c]{22.17} & \makecell[c]{17.37} & \makecell[c]{21.20} & \makecell[c]{0.859} & \makecell[c]{0.766} & \makecell[c]{0.549} & \makecell[c]{0.707}\\
                \hline
                \makecell[c]{SIFT\cite{lowe2004distinctive}+MAGSAC\cite{barath2019magsac}} & \makecell[c]{25.51} & \makecell[c]{22.21} & \makecell[c]{17.14} & \makecell[c]{21.17} & \makecell[c]{0.868} & \makecell[c]{0.767} & \makecell[c]{0.544} & \makecell[c]{0.708}\\
                \hline
                \makecell[c]{ORB\cite{rublee2011orb}+MAGSAC\cite{barath2019magsac}} & \makecell[c]{22.33} & \makecell[c]{18.56} & \makecell[c]{13.71} & \makecell[c]{17.75} & \makecell[c]{0.754} & \makecell[c]{0.587} & \makecell[c]{0.369} & \makecell[c]{0.550}\\
                \hline
                \makecell[c]{BRISK\cite{leutenegger2011brisk}+MAGSAC\cite{barath2019magsac}} & \makecell[c]{24.18} & \makecell[c]{21.04} & \makecell[c]{16.04} & \makecell[c]{19.98} & \makecell[c]{0.822} & \makecell[c]{0.714} & \makecell[c]{0.498} & \makecell[c]{0.660}\\
                \hline
                \makecell[c]{SOSNet\cite{sosnet2019cvpr}+MAGSAC\cite{barath2019magsac}} & \makecell[c]{25.42} & \makecell[c]{21.75} & \makecell[c]{16.10} & \makecell[c]{20.59} & \makecell[c]{0.864} & \makecell[c]{0.747} & \makecell[c]{0.495} & \makecell[c]{0.681}\\
                \hline
                \makecell[c]{SIFT\cite{lowe2004distinctive}+MAGSAC++\cite{barath2019magsacplusplus}} & \makecell[c]{25.64} & \makecell[c]{22.32} & \makecell[c]{17.01} & \makecell[c]{21.19} & \makecell[c]{0.870} & \makecell[c]{0.772} & \makecell[c]{0.539} & \makecell[c]{0.708}\\
                \hline
                \makecell[c]{ORB\cite{rublee2011orb}+MAGSAC++\cite{barath2019magsacplusplus}} & \makecell[c]{22.36} & \makecell[c]{18.86} & \makecell[c]{13.56} & \makecell[c]{17.79} & \makecell[c]{0.755} & \makecell[c]{0.595} & \makecell[c]{0.370} & \makecell[c]{0.553}\\
                \hline
                \makecell[c]{BRISK\cite{leutenegger2011brisk}+MAGSAC++\cite{barath2019magsacplusplus}} & \makecell[c]{24.63} & \makecell[c]{21.41} & \makecell[c]{16.17} & \makecell[c]{20.28} & \makecell[c]{0.839} & \makecell[c]{0.730} & \makecell[c]{0.501} & \makecell[c]{0.671}\\
                \hline
                \makecell[c]{SOSNet\cite{sosnet2019cvpr}+MAGSAC++\cite{barath2019magsacplusplus}} & \makecell[c]{25.59} & \makecell[c]{21.98} & \makecell[c]{16.15} & \makecell[c]{20.72} & \makecell[c]{0.869} & \makecell[c]{0.761} & \makecell[c]{0.499} & \makecell[c]{0.689}\\
                \hline
                \hline
                \makecell[c]{DHN\cite{detone2016deep}} & \makecell[c]{16.40} & \makecell[c]{13.36} & \makecell[c]{11.48} & \makecell[c]{13.52} & \makecell[c]{0.409} & \makecell[c]{0.170} & \makecell[c]{0.076} & \makecell[c]{0.204} \\
                \hline
                \makecell[c]{UDHN\cite{nguyen2018unsupervised}} & \makecell[c]{19.39} & \makecell[c]{15.93} & \makecell[c]{13.09} & \makecell[c]{15.83} & \makecell[c]{0.573} & \makecell[c]{0.334} & \makecell[c]{0.165} & \makecell[c]{0.338}\\
                \hline
                \makecell[c]{CA-DHN\cite{zhang2019content}} & \makecell[c]{$F$} & \makecell[c]{13.13} & \makecell[c]{11.00} & \makecell[c]{13.16} & \makecell[c]{0.339} & \makecell[c]{0.181} & \makecell[c]{0.105} & \makecell[c]{0.198} \\
                \hline
                \makecell[c]{LB-DHN\cite{nie2020learning}} & \makecell[c]{24.75} & \makecell[c]{21.14} & \makecell[c]{18.43} & \makecell[c]{21.14} & \makecell[c]{0.825} & \makecell[c]{0.712} & \makecell[c]{0.547} & \makecell[c]{0.680} \\
                \hline
                \makecell[c]{LB-UDHN\cite{9472883}} & \makecell[c]{$\mathbf{\textcolor{blue}{27.84}}$} & \makecell[c]{$\mathbf{\textcolor{blue}{23.95}}$} & \makecell[c]{$\mathbf{\textcolor{blue}{20.70}}$} & \makecell[c]{$\mathbf{\textcolor{blue}{23.80}}$} & \makecell[c]{$\mathbf{\textcolor{blue}{0.902}}$} & \makecell[c]{$\mathbf{\textcolor{blue}{0.830}}$} & \makecell[c]{$\mathbf{\textcolor{blue}{0.685}}$} & \makecell[c]{$\mathbf{\textcolor{blue}{0.793}}$} \\
                \hline
                \hline
                \makecell[c]{Ours ($8\times 8$)} & \makecell[c]{$\mathbf{\textcolor{red}{28.41}}$} & \makecell[c]{$\mathbf{\textcolor{red}{24.63}}$} & \makecell[c]{$\mathbf{\textcolor{red}{21.59}}$} & \makecell[c]{$\mathbf{\textcolor{red}{24.54}}$} & \makecell[c]{$\mathbf{\textcolor{red}{0.913}}$} & \makecell[c]{$\mathbf{\textcolor{red}{0.853}}$} & \makecell[c]{$\mathbf{\textcolor{red}{0.733}}$} & \makecell[c]{$\mathbf{\textcolor{red}{0.823}}$}\\
                \hline

          \end{tabular}
        }
    \label{fistTable}
}

\qquad

\subtable[Quantitative comparison on original resolution.]{
    \scalebox{0.90}{
    \begin{tabular}{|l||l|l|l|l||l|l|l|l|}
        \hline
        \makecell[c]{\multirow{3}{*}{}}  & \multicolumn{8}{c|}{$512\times 512$} \\
        \cline{2-9}
        & \multicolumn{4}{c||}{PSNR ($\uparrow$)} & \multicolumn{4}{c|}{SSIM ($\uparrow$)} \\
        \cline{2-9}
        & \makecell[c]{Easy} & Moderate & Hard & \makecell[c]{Average} & \makecell[c]{Easy} & Moderate & Hard & \makecell[c]{Average} \\
        \hline
        \hline
        \makecell[c]{$I_{3\times 3}$} & \makecell[c]{15.87} & \makecell[c]{12.76} & \makecell[c]{10.68} & \makecell[c]{12.86} & \makecell[c]{0.530} & \makecell[c]{0.286} & \makecell[c]{0.146} & \makecell[c]{0.303}\\
        \hline
        \hline
        \makecell[c]{SIFT\cite{lowe2004distinctive}+RANSAC\cite{fischler1981random}} & \makecell[c]{28.75} & \makecell[c]{24.08} & \makecell[c]{18.55} & \makecell[c]{23.27} & \makecell[c]{0.916} & \makecell[c]{0.833} & \makecell[c]{0.636} & \makecell[c]{0.779} \\
        \hline
        \makecell[c]{ORB\cite{rublee2011orb}+RANSAC\cite{fischler1981random}}& \makecell[c]{27.53} & \makecell[c]{22.85} & \makecell[c]{17.37} & \makecell[c]{22.06} & \makecell[c]{0.888} & \makecell[c]{0.772} & \makecell[c]{0.550} & \makecell[c]{0.718}\\
        \hline
        \makecell[c]{BRISK\cite{leutenegger2011brisk}+RANSAC\cite{fischler1981random}} & \makecell[c]{28.57} & \makecell[c]{24.02} & \makecell[c]{18.63} & \makecell[c]{23.23} & \makecell[c]{0.911} & \makecell[c]{0.824} & \makecell[c]{0.629} & \makecell[c]{0.772}\\
        \hline
        \makecell[c]{SOSNet\cite{sosnet2019cvpr}+RANSAC\cite{fischler1981random}} & \makecell[c]{28.60} & \makecell[c]{23.85} & \makecell[c]{17.47} & \makecell[c]{22.71} & \makecell[c]{0.915} & \makecell[c]{0.824} & \makecell[c]{0.596} & \makecell[c]{0.760}\\
        \hline
        \makecell[c]{SIFT\cite{lowe2004distinctive}+MAGSAC\cite{barath2019magsac}}  & \makecell[c]{27.42} & \makecell[c]{22.39} & \makecell[c]{17.07} & \makecell[c]{21.77} & \makecell[c]{0.891} & \makecell[c]{0.777} & \makecell[c]{0.574} & \makecell[c]{0.730}\\
        \hline
        \makecell[c]{ORB\cite{rublee2011orb}+MAGSAC\cite{barath2019magsac}} & \makecell[c]{26.36} & \makecell[c]{21.49} & \makecell[c]{16.26} & \makecell[c]{20.86} & \makecell[c]{0.867} & \makecell[c]{0.729} & \makecell[c]{0.516} & \makecell[c]{0.685}\\
        \hline
        \makecell[c]{BRISK\cite{leutenegger2011brisk}+MAGSAC\cite{barath2019magsac}} & \makecell[c]{26.28} & \makecell[c]{20.82} & \makecell[c]{16.25} & \makecell[c]{20.63} & \makecell[c]{0.868} & \makecell[c]{0.710} & \makecell[c]{0.504} & \makecell[c]{0.675}\\
        \hline
        \makecell[c]{SOSNet\cite{sosnet2019cvpr}+MAGSAC\cite{barath2019magsac}} & \makecell[c]{26.13} & \makecell[c]{19.55} & \makecell[c]{14.25} & \makecell[c]{19.39} & \makecell[c]{0.867} & \makecell[c]{0.693} & \makecell[c]{0.456} & \makecell[c]{0.650}\\
        \hline
        \makecell[c]{SIFT\cite{lowe2004distinctive}+MAGSAC++\cite{barath2019magsacplusplus}} & \makecell[c]{27.90} & \makecell[c]{22.39} & \makecell[c]{17.11} & \makecell[c]{21.93} & \makecell[c]{0.901} & \makecell[c]{0.779} & \makecell[c]{0.565} & \makecell[c]{0.730}\\
        \hline
        \makecell[c]{ORB\cite{rublee2011orb}+MAGSAC++\cite{barath2019magsacplusplus}} & \makecell[c]{27.32} & \makecell[c]{22.02} & \makecell[c]{16.05} & \makecell[c]{21.22} & \makecell[c]{0.884} & \makecell[c]{0.749} & \makecell[c]{0.515} & \makecell[c]{0.696}\\
        \hline
        \makecell[c]{BRISK\cite{leutenegger2011brisk}+MAGSAC++\cite{barath2019magsacplusplus}} & \makecell[c]{27.05} & \makecell[c]{20.89} & \makecell[c]{16.02} & \makecell[c]{20.79} & \makecell[c]{0.884} & \makecell[c]{0.717} & \makecell[c]{0.497} & \makecell[c]{0.679}\\
        \hline
        \makecell[c]{SOSNet\cite{sosnet2019cvpr}+MAGSAC++\cite{barath2019magsacplusplus}}& \makecell[c]{26.33} & \makecell[c]{19.26} & \makecell[c]{13.73} & \makecell[c]{19.16} & \makecell[c]{0.872} & \makecell[c]{0.675} & \makecell[c]{0.433} & \makecell[c]{0.637}\\
        \hline
        \hline
        \makecell[c]{DHN\cite{detone2016deep}} & \makecell[c]{-} & \makecell[c]{-} & \makecell[c]{-} & \makecell[c]{-} & \makecell[c]{-} & \makecell[c]{-} & \makecell[c]{-} & \makecell[c]{-}\\
        \hline
        \makecell[c]{UDHN\cite{nguyen2018unsupervised}} & \makecell[c]{-} & \makecell[c]{-} & \makecell[c]{-} & \makecell[c]{-} & \makecell[c]{-} & \makecell[c]{-} & \makecell[c]{-} & \makecell[c]{-}\\
        \hline
        \makecell[c]{CA-DHN\cite{zhang2019content}} & \makecell[c]{$F$} & \makecell[c]{$F$} & \makecell[c]{$F$} & \makecell[c]{$F$} & \makecell[c]{$F$} & \makecell[c]{$F$} & \makecell[c]{0.150} & \makecell[c]{$F$}\\
        \hline
        \makecell[c]{LB-DHN\cite{nie2020learning}} & \makecell[c]{-} & \makecell[c]{-} & \makecell[c]{-} & \makecell[c]{-} & \makecell[c]{-} & \makecell[c]{-} & \makecell[c]{-} & \makecell[c]{-}\\
        \hline
        \makecell[c]{LB-UDHN\cite{9472883}} & \makecell[c]{-} & \makecell[c]{-} & \makecell[c]{-} & \makecell[c]{-} & \makecell[c]{-} & \makecell[c]{-} & \makecell[c]{-} & \makecell[c]{-}\\
        \hline
        \hline
        \makecell[c]{APAP\cite{zaragoza2013projective} ($8\times 8$)} &\makecell[c]{27.58} & \makecell[c]{23.92} & \makecell[c]{19.90} & \makecell[c]{23.41} & \makecell[c]{0.895} & \makecell[c]{0.824} & \makecell[c]{0.663} & \makecell[c]{0.781}\\
        \hline
        \makecell[c]{AANAP\cite{lin2015adaptive} ($8\times 8$)} & \makecell[c]{27.59} & \makecell[c]{23.85} & \makecell[c]{19.70} & \makecell[c]{23.31} & \makecell[c]{0.896} & \makecell[c]{0.823} & \makecell[c]{0.659} & \makecell[c]{0.779}\\
        \hline
        \makecell[c]{robust ELA\cite{li2017parallax} ($8\times 8$)} & \makecell[c]{28.08} & \makecell[c]{23.62} & \makecell[c]{18.16} & \makecell[c]{22.77} & \makecell[c]{0.896} & \makecell[c]{0.803} & \makecell[c]{0.620} & \makecell[c]{0.758}\\
        \hline
        \makecell[c]{SPW\cite{liao2019single} ($8\times 8$)} & \makecell[c]{26.76} & \makecell[c]{22.48} & \makecell[c]{15.92} & \makecell[c]{21.13} & \makecell[c]{0.875} & \makecell[c]{0.748} & \makecell[c]{0.442} & \makecell[c]{0.663}\\
        \hline
        \makecell[c]{LCP\cite{jia2021leveraging} ($8\times 8$)} & \makecell[c]{26.96} & \makecell[c]{22.59} & \makecell[c]{19.24} & \makecell[c]{22.56} & \makecell[c]{0.878} & \makecell[c]{0.763} & \makecell[c]{0.606} & \makecell[c]{0.734}\\
        \hline

        \hline
        \makecell[c]{APAP\cite{zaragoza2013projective} ($100\times 100$)} &\makecell[c]{27.96} & \makecell[c]{24.39} & \makecell[c]{$\mathbf{\textcolor{blue}{20.21}}$} & \makecell[c]{23.79} & \makecell[c]{0.901} & \makecell[c]{0.837} & \makecell[c]{0.682} & \makecell[c]{0.794}\\
        \hline
        \makecell[c]{AANAP\cite{lin2015adaptive} ($100\times 100$)} & \makecell[c]{27.80} & \makecell[c]{24.05} & \makecell[c]{19.91} & \makecell[c]{23.52} & \makecell[c]{0.897} & \makecell[c]{0.829} & \makecell[c]{0.668} & \makecell[c]{0.785}\\
        \hline
        \makecell[c]{robust ELA\cite{li2017parallax} ($100\times 100$)} & \makecell[c]{$\mathbf{\textcolor{blue}{29.28}}$} & \makecell[c]{$\mathbf{\textcolor{blue}{25.12}}$} & \makecell[c]{19.15} & \makecell[c]{$\mathbf{\textcolor{blue}{23.98}}$} & \makecell[c]{$\mathbf{\textcolor{blue}{0.917}}$} & \makecell[c]{$\mathbf{\textcolor{blue}{0.853}}$} & \makecell[c]{$\mathbf{\textcolor{blue}{0.689}}$} & \makecell[c]{$\mathbf{\textcolor{blue}{0.806}}$}\\
        \hline
        \makecell[c]{SPW\cite{liao2019single} ($100\times 100$)} & \makecell[c]{26.87} & \makecell[c]{22.74} & \makecell[c]{16.85} & \makecell[c]{21.61} & \makecell[c]{0.880} & \makecell[c]{0.759} & \makecell[c]{0.494} & \makecell[c]{0.689}\\
        \hline
        \makecell[c]{LCP\cite{jia2021leveraging} ($100\times 100$)} & \makecell[c]{26.43} & \makecell[c]{22.53} & \makecell[c]{19.28} & \makecell[c]{22.40} & \makecell[c]{0.873} & \makecell[c]{0.760} & \makecell[c]{0.611} & \makecell[c]{0.734}\\
        \hline
        \hline
        \makecell[c]{Ours ($8\times 8$)} & \makecell[c]{$\mathbf{\textcolor{red}{29.52}}$} & \makecell[c]{$\mathbf{\textcolor{red}{25.24}}$} & \makecell[c]{$\mathbf{\textcolor{red}{21.20}}$} & \makecell[c]{$\mathbf{\textcolor{red}{24.89}}$} & \makecell[c]{$\mathbf{\textcolor{red}{0.923}}$} & \makecell[c]{$\mathbf{\textcolor{red}{0.859}}$} & \makecell[c]{$\mathbf{\textcolor{red}{0.708}}$} & \makecell[c]{$\mathbf{\textcolor{red}{0.817}}$}\\
        \hline

  \end{tabular}
}
\label{secondTable}
}
\label{UDIS-D_TABLE}
\end{table*}

\begin{figure*}[!t]
    \centering
    \includegraphics[width=1\textwidth]{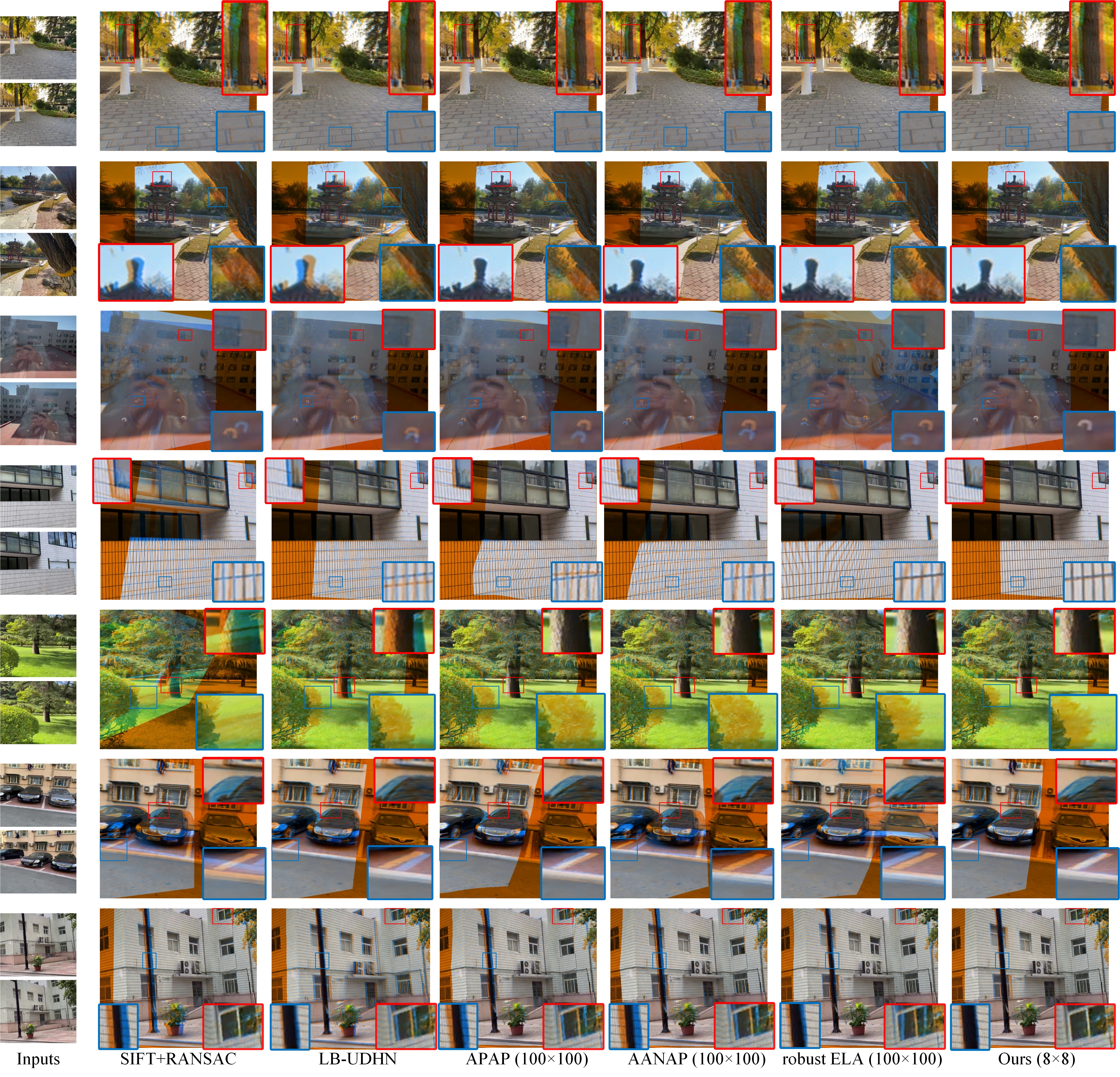}
    \vspace{-0.8cm}
    \caption{Qualitative comparison with the state-of-the-arts on real-world dataset (UDIS-D). For each instance, we magnify a far region in a red box and a near region in a blue box to compare the alignment performance in scenes with parallax. The mesh-based methods adopt different mesh sizes because they reach the best performance at that size. (The details can be found in Table \ref{secondTable} and Table \ref{ablation}.)}
    \label{Qualitative_comparison}
\end{figure*}

\subsection{Comparison on Real-World Dataset}
\label{section43}

\textbf{Quantitative Comparison.}

The real-world image pairs that contain rich depth levels are captured under different camera baselines, which indicates they can not be aligned using a single homography due to the parallax. Therefore, in this real-world dataset (UDIS-D), we add a comparison with the multi-homography solutions --- APAP \cite{zaragoza2013projective}, AANAP \cite{lin2015adaptive}, robust ELA \cite{li2017parallax}, SPW \cite{liao2019single} and LCP \cite{jia2021leveraging}. Since the mesh in our network is $8\times 8$ (the reason for adopting $8\times 8$ mesh in our network will be explained in Section \ref{section45}), it is fair to compare ours with these methods that set the mesh to $8\times 8$. To further highlight our performance, we also compare them with the mesh set to $100\times 100$.

In addition, we carry on the experiments in UDIS-D in two different resolutions as shown in Table \ref{UDIS-D_TABLE}. Formally, the resolution of images in UDIS-D is $512\times 512$, we resize them to $128\times 128$ to get a low-resolution version for two reasons: 1) The low-resolution images can simulate the challenging scenes with fewer feature points. 2) Most existing deep learning solutions can only work at a fixed resolution ($128\times 128$) due to the existence of fully connected layers. Following \cite{9472883}, we evaluate the performance using PSNR and SSIM in the overlapping regions, which can be calculated as follows:
\begin{equation}
    \begin{aligned}
    PSNR_{overlap} = \mathcal{PSNR}\ (\mathcal{W}(E)\odot I_r , \mathcal{W}(I_t)), \\
    SSIM_{overlap} = \mathcal{SSIM}\ (\mathcal{W}(E)\odot I_r , \mathcal{W}(I_t)),
    \end{aligned}
    \label{EQ_ablation-based}
\end{equation}
where $\mathcal{PSNR}(\cdot)$ and $\mathcal{SSIM}(\cdot)$ denote the operations of computing PSNR and SSIM between two images, respectively.

Form Table \ref{UDIS-D_TABLE}, we can observe that:

(1) Our method is better than all the homography estimation methods in each case ( ``easy'', ``moderate'' and ``hard'') of different resolutions (128$\times$128 and 512$\times$512).

(2) The performance of the traditional solutions dorp as the resolution decreases because the performance of feature detection degrades in the low-resolution scenes.

(3) Most learning-based homography solutions do not work well because the overlap rates between the image pairs in this dataset are low. The long-range correlation can not be effectively learned by the convolutional layers or cost volume.

(4) The latest traditional multi-homography solutions (SPW and LCP) show unsatisfying performance in UDIS-D. The main reason is that they introduce other constraints except for the point-line alignment term such as distortion term and line preserving term, reducing projective distortions at the cost of sacrificing alignment performance. Meanwhile, the point-line alignment could not work well in every scene, especially where the matched points and lines are not sufficient.

\begin{spacing}{1.5}
\end{spacing}
\textbf{Qualitative Comparison.}

In addition to Table \ref{UDIS-D_TABLE}, we also provide a qualitative comparison between these methods in the original resolution of this real-world dataset. We choose the combination of SIFT and RANSAC as the representative work of traditional single homography solutions since it achieves the best alignment performance in UDIS-D.
Also, LB-UDHN is chosen as the representative work of deep learning-based solutions for the best performance. To apply it to images with higher resolution (512$\times$512), we multiply the estimated homography matrix by a scale transformation matrix to adapt to other resolutions.
As for multi-homography solutions, we demonstrate their results with the mesh sets to $100\times 100$. Eight instances are displayed in Fig. \ref{Qualitative_comparison}, where each instance contains varying degrees of parallax. We fuse the reference image and the warped target image by setting the intensity of the blue channel in the reference image and that of the red channel in the warped target image to zero. In this manner, the non-overlapping regions are shown in orange, and the misalignments in the overlapping regions are highlighted in a different color. Although the proposed method cannot remove all the misalignments, the remaining misalignments in our results are less than that of other methods.

\begin{figure}[!t]
    \centering
    \includegraphics[width=0.47\textwidth]{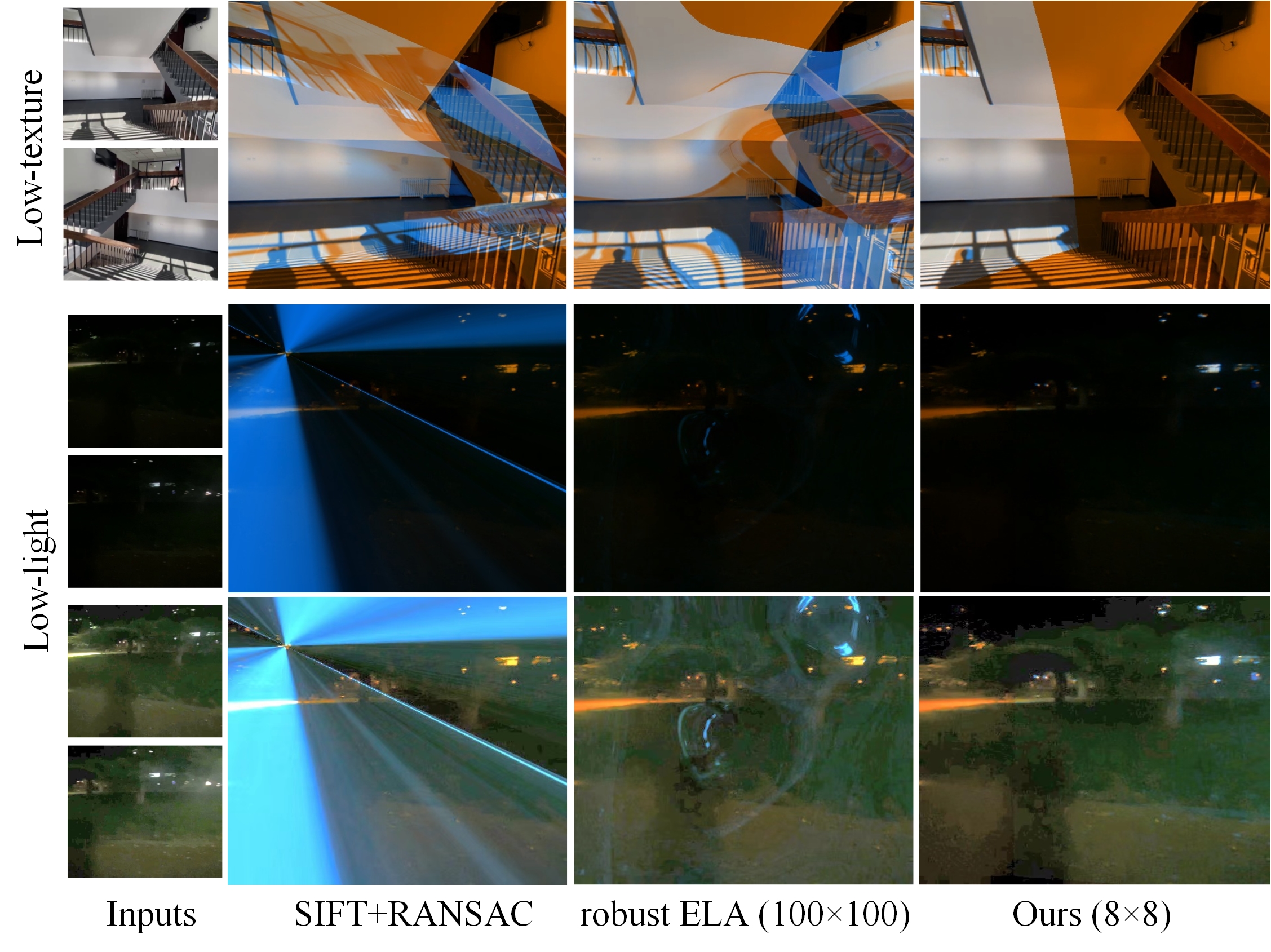}
    \vspace{-0.4cm}
    \caption{Robust analysis on challenging cases. Row 1: A low-texture scene. Row 2-3: A low-light scene. The example in Row 2 is too challenging that we apply image augmentation on them, and the augmented results are demonstrated in Row 3.}
    \label{Robustness_Analysis}
\end{figure}

\begin{spacing}{1.5}
\end{spacing}
\textbf{Robustness Analysis.}

 A robust method does not necessarily require good average performance (see `average' columns of Table \ref{UDIS-D_TABLE}) but requires that the worst performance cannot be poor (see 'hard' columns of Table \ref{UDIS-D_TABLE}).
 Comparing Table \ref{fistTable} with Table \ref{secondTable}, the performance of traditional solutions in high-resolution is significantly better because the feature points are more abundant when the resolution is increased.

 However, for some challenging scenes such as low-light or low-texture, the increase in resolution does not improve the performance. Fig. \ref{Robustness_Analysis} demonstrates two challenging examples. When the texture is missing or the light is too low in a scene, the manual designed feature descriptors, $e.g.$ SIFT, ORB and $etc.$, are not applicable, resulting in few feature points or mismatched feature correspondences. Compared with them, the deep learning solutions extract the feature maps by learning the distribution of dataset samples adaptively. Hence, the learning solutions are more robust.

\begin{figure*}[!t]
    \centering
    \subfigure[Railtracks \cite{zaragoza2013projective}.]
    {\includegraphics[width=0.18\textwidth]{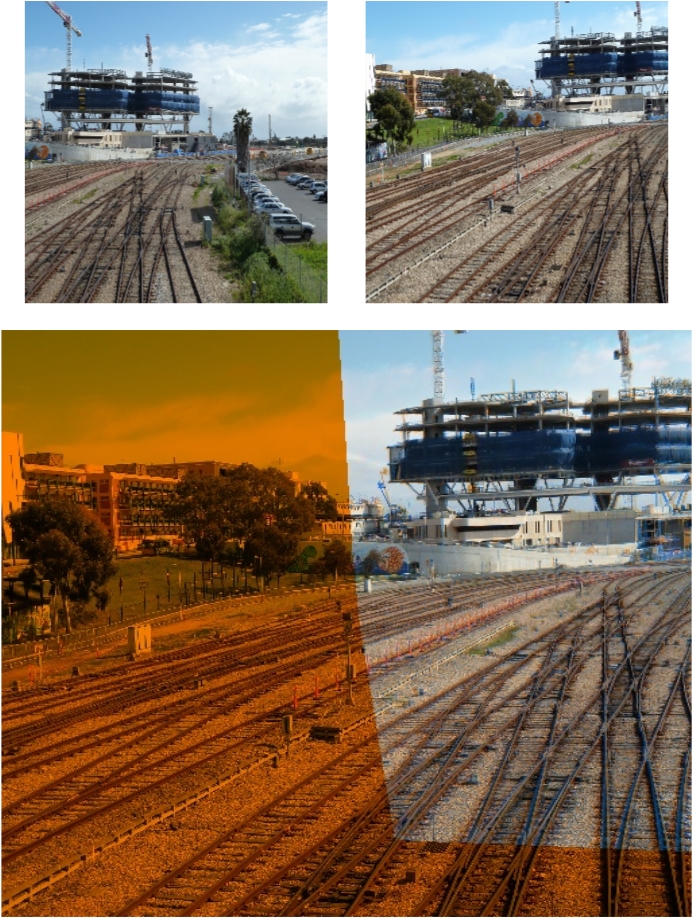}}
    \subfigure[Yard \cite{gao2013seam}.]
    {\includegraphics[width=0.18\textwidth]{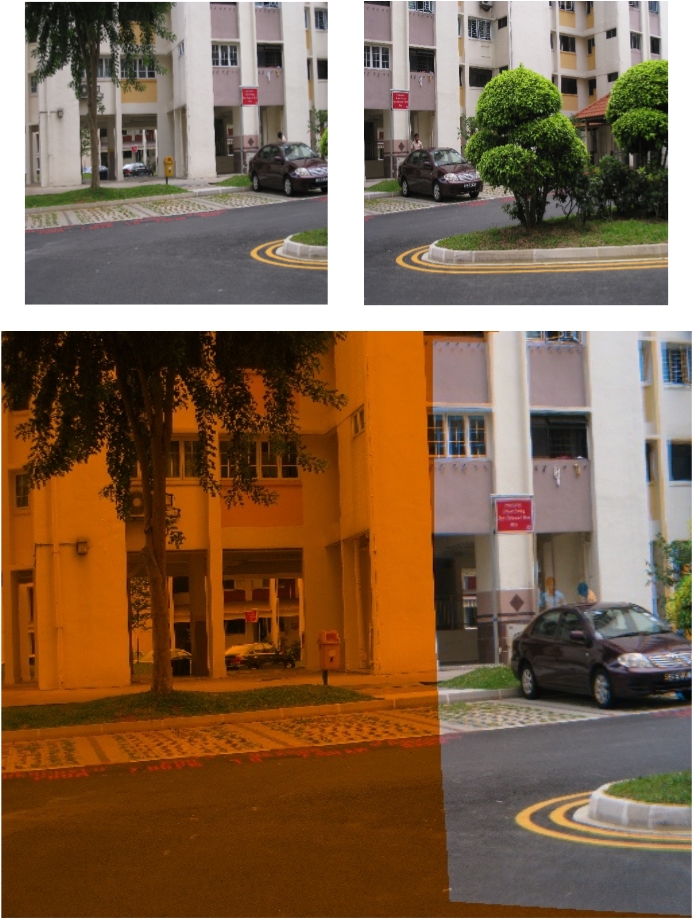}}
    \subfigure[Carpark \cite{chang2014shape}.]
    {\includegraphics[width=0.18\textwidth]{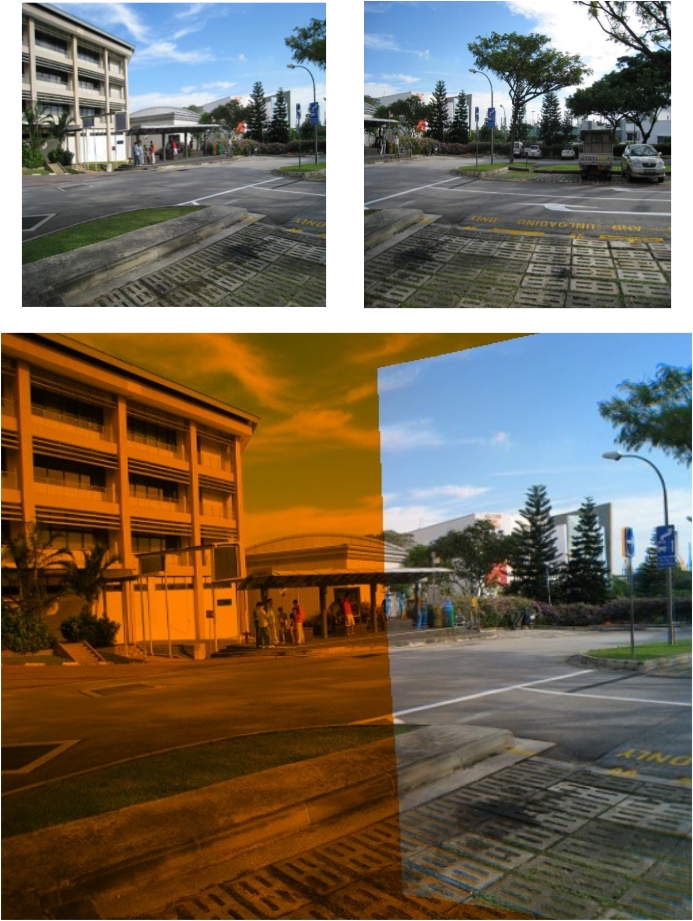}}
    \subfigure[Temple \cite{gao2011constructing}.]
    {\includegraphics[width=0.18\textwidth]{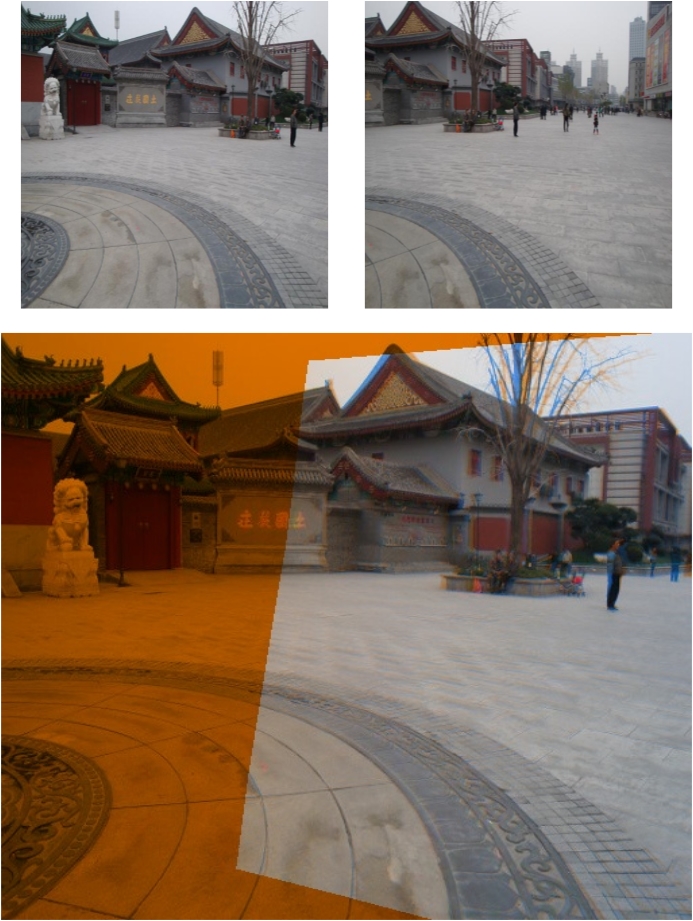}}
    \subfigure[Garden \cite{li2019local}.]
    {\includegraphics[width=0.18\textwidth]{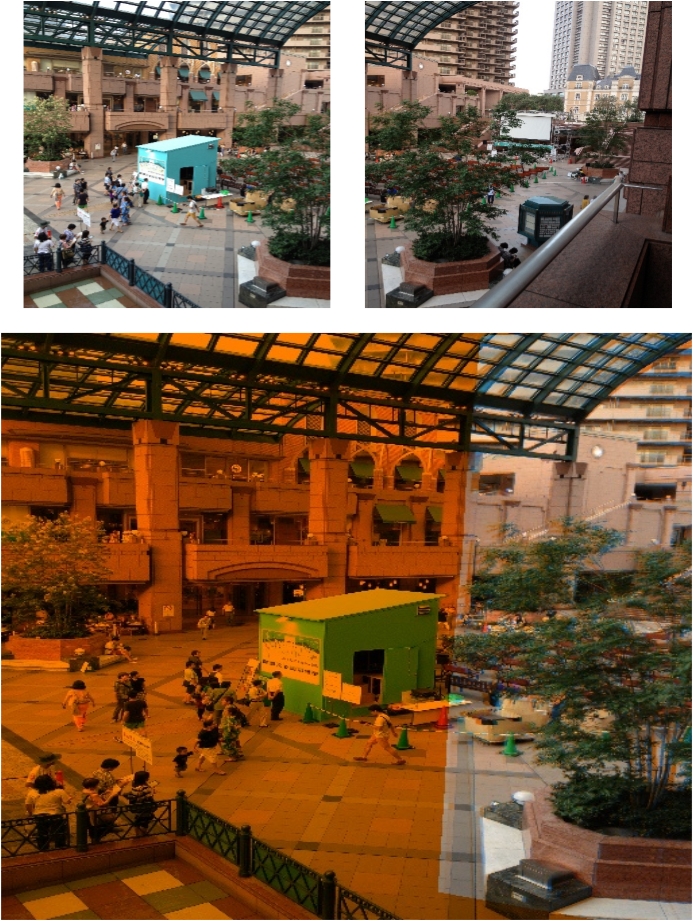}}
    \vspace{-0.2cm}
    \caption{Cross-dataset validation on generalization ability. Top to bottom: inputs and our aligned results.}
    \label{cross_dataset}
 \end{figure*}

\begin{spacing}{1.5}
\end{spacing}
\textbf{Cross-dataset Validattion.}

For most deep learning methods, generalization is a flaw. In this section, we validate the generalization ability of the proposed method across datasets. Specifically, we train our network in UDIS-D and test it in other datasets. We collect the datasets from classic image stitching papers \cite{zaragoza2013projective, gao2013seam, chang2014shape, gao2011constructing, li2019local}, where these datasets are captured from different scenes and contain various degrees of parallax. The results are shown in Fig. \ref{cross_dataset}. Even if it is tested on other datasets, our solution still has good alignment capabilities. Especially the overlap rates between input images in these examples are pretty low and most deep learning solutions could fail in these scenes.

\begin{figure}[!t]
    \centering
    \includegraphics[width=0.47\textwidth]{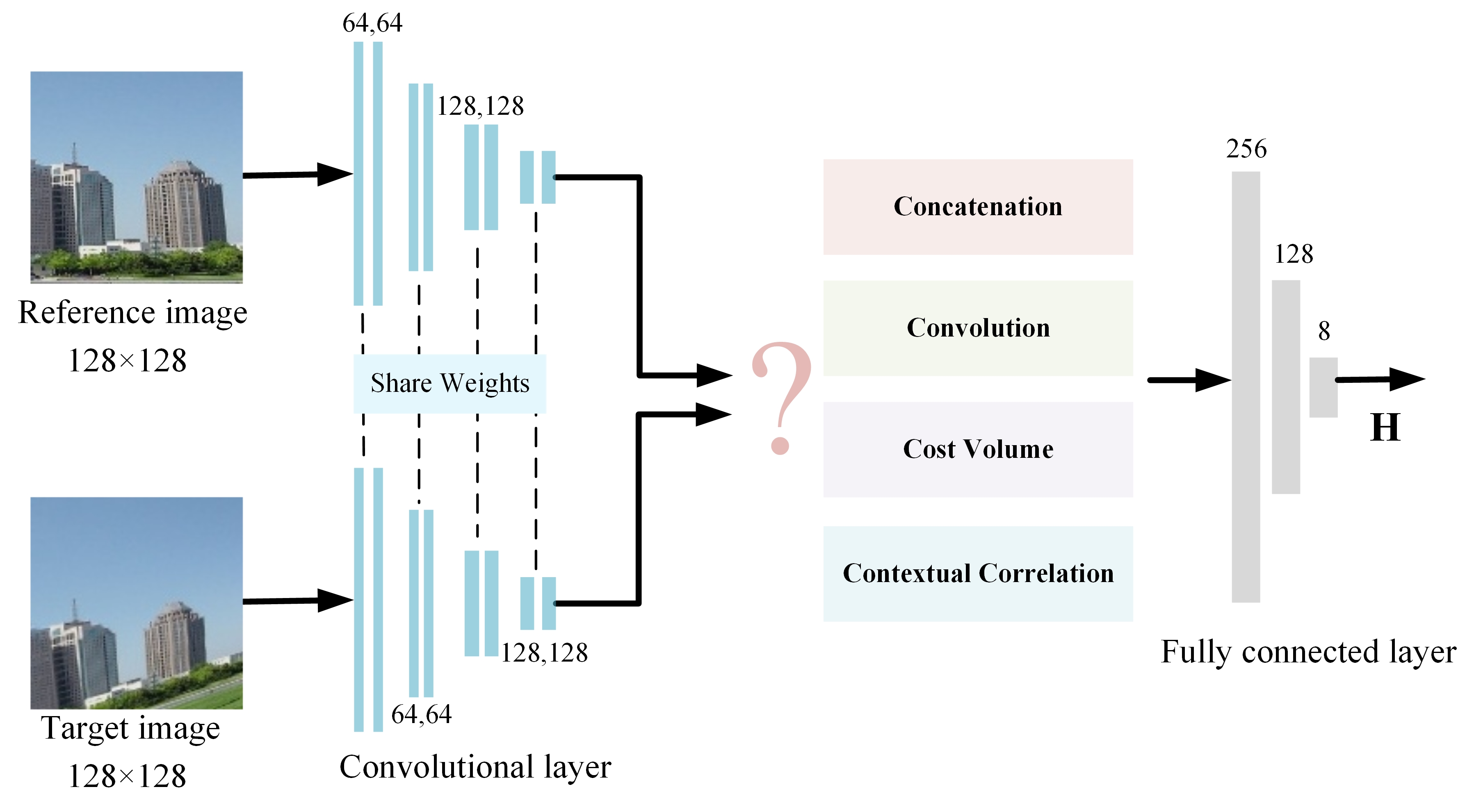}
    \vspace{-0.4cm}
    \caption{The network architecture of the deep homography baseline in Section \ref{section44}.}
    \label{baseline}
\end{figure}

\begin{table*}[]
    \centering
    \caption{Comprehensive comparison of the performance, the number of parameters, and the speed between the cost volume and proposed CCL on Warped MS-COCO.}

\scalebox{0.9}{
    \begin{tabular}{|l||l|l|l|}
        \hline
        & \makecell[c]{Performance ($\downarrow$)\\(4-pt homography RMSE)}& \makecell[c]{Model size ($\downarrow$) \\(/MB)}& \makecell[c]{Time per iteration ($\downarrow$) \\(/ms)}\\
        \hline
        \hline
        \makecell[c]{Concatenation} & \makecell[c]{18.5218}& \makecell[c]{199.58}& \makecell[c]{\textbf{6.3500}}\\
        \hline
        \makecell[c]{Convolution} & \makecell[c]{18.5218}& \makecell[c]{213.08}& \makecell[c]{6.6372}\\
        \hline
        \makecell[c]{Cost Volume} & \makecell[c]{14.0783}& \makecell[c]{824.33}& \makecell[c]{40.0687}\\
        \hline
        \makecell[c]{CCL} & \makecell[c]{\textbf{4.7752}}& \makecell[c]{\textbf{9.08}}& \makecell[c]{6.9576}\\
        \hline

  \end{tabular}
}
\label{correlation_compare}
\end{table*}

\subsection{Compared with Cost Volume}
\label{section44}

In addition to image alignment performance, we also compare the proposed CCL with the cost volume in the field of homography estimation. Both CCL and cost volume are modules used to extract matching information between feature correspondences without learning parameters. Both of them can easily be plugged into a neural network.

At first, we replaced the CCL with the cost volume in our network. Nevertheless, this replacement will bring about a dramatic increase in the number of network parameters, causing the training crash by exceeding the maximum memory of the GPU. Therefore, we design a relatively simple network as the baseline and compare the proposed CCL with the cost volume on this framework. The baseline architecture is shown in Fig. \ref{baseline}, where 8 convolutional layers with shared parameters and 3 max-pooling layers are adopted to extract the deep features. Then the matching information can be obtained from the feature maps extracted from different input images using a certain approach. Specifically, directly concatenating two convolutional layers with filter number setting to 256, the cost volume with search radius setting to 16, and the proposed CCL are adopted in different experiments. After that, we use three fully connected layers to predict the 8 motions of the 4 vertices in the target image.

The experiments are conducted on Warped MS-COCO in one RTX NVIDIA 2080 Ti and all the schemes are trained in a supervised manner for 80 epochs. The results are shown in Table \ref{correlation_compare}, where a comprehensive evaluation of the performance, the number of parameters, and the speed is formulated. From this table, we can observe:

(1) The performance of ``Concatenation'' is almost equal to $I_{3\times 3}$ (18.5220, as shown in Table \ref{synthetic_dataset}). Without the matching relationship between the input feature maps, the network learns nothing.

(2) The performance of ``Convolution'' is almost the same as that of ``Concatenation'', which indicates convolutional layers learn nothing about the matching relationship. The limited receptive field can interpret this phenomenon.

(3) The cost volume does help to extract the matching relationship between the input feature maps, but it is not efficient. Besides, the costs on parameters and speed are significantly increased.

(4) The performance of the proposed CCL is much better than that of the cost volume. In addition, this efficient module only increases 0.32$ms$ running time compared with ``Convolution''. As for the model size, our design significantly decreases the parameters from several hundred MB to 10 MB.

Actually, the decrease in the model size mainly owes to the feature flow. This lightweight representation of feature correlation ($H\times W\times2$) significantly reduces the parameters of the subsequent fully connected layer. Compared with the cost volume, the feature flow also rejects extensive redundant matching information, leading to more efficient prediction on homography.

\begin{table}[]
    \centering
    \caption{Ablation studies of contextual correlation on Warped MS-COCO.}

\scalebox{0.9}{
    \begin{tabular}{|l||l|l|l|l|}

        \hline
        \makecell[c]{Correlation Volume} & \makecell[c]{\CheckmarkBold}&  \makecell[c]{\CheckmarkBold}& \makecell[c]{\CheckmarkBold}\\
        \hline
        \makecell[c]{Scale Softmax} & & & \makecell[c]{\CheckmarkBold}\\
        \hline
        \makecell[c]{Feature Flow} & & \makecell[c]{\CheckmarkBold}& \makecell[c]{\CheckmarkBold}\\
        \hline
        \hline
        \makecell[c]{Performance ($\downarrow$) \\(4-pt homography RMSE)} &\makecell[c]{18.5218} &\makecell[c]{6.6469} &\makecell[c]{4.7752} \\
        \hline
        \makecell[c]{Model size ($\downarrow$) (/MB)} &\makecell[c]{199.58} &\makecell[c]{9.08} &\makecell[c]{9.08} \\
        \hline
        \makecell[c]{Time per iteration ($\downarrow$) (/ms)} &\makecell[c]{7.2594} &\makecell[c]{6.9534} &\makecell[c]{6.9576} \\
        \hline

  \end{tabular}
}
\label{ablation_corraltion}
\end{table}

\begin{table}[]
    \centering
    \caption{Ablation studies of the number of grids and depth-aware shape-preserved loss on UDIS-D.}

\scalebox{0.75}{
    \begin{tabular}{|l|l||l|l|}
        \hline
        Multi-Grid & Depth-Aware SPL& \makecell[c]{Average PSNR ($\uparrow$)}& \makecell[c]{Average SSIM ($\uparrow$)}\\
        \hline
        \hline
        \makecell[c]{1$\times$1} &\makecell[c]{\XSolidBrush}& \makecell[c]{23.1144}& \makecell[c]{0.7541}\\
        \hline
        \makecell[c]{2$\times$2} &\makecell[c]{\XSolidBrush}& \makecell[c]{23.1885}& \makecell[c]{0.7598}\\
        \hline
        \makecell[c]{4$\times$4} &\makecell[c]{\XSolidBrush}& \makecell[c]{24.6234}& \makecell[c]{0.8093}\\
        \hline
        \makecell[c]{8$\times$8} &\makecell[c]{\XSolidBrush}& \makecell[c]{\textbf{24.7581}}& \makecell[c]{\textbf{0.8129}}\\
        \hline
        \makecell[c]{16$\times$16} &\makecell[c]{\XSolidBrush}& \makecell[c]{24.6360}& \makecell[c]{0.8086}\\
        \hline
        \hline
        \makecell[c]{8$\times$8} & \makecell[c]{1 depth levels} & \makecell[c]{24.7124}& \makecell[c]{0.8105}\\
        \hline
        \makecell[c]{8$\times$8} & \makecell[c]{2 depth levels} & \makecell[c]{24.7262}& \makecell[c]{0.8113}\\
        \hline
        \makecell[c]{8$\times$8} & \makecell[c]{4 depth levels}& \makecell[c]{24.7669}& \makecell[c]{0.8134}\\
        \hline
        \makecell[c]{8$\times$8} & \makecell[c]{8 depth levels}& \makecell[c]{24.8145}& \makecell[c]{0.8146}\\
        \hline
        \makecell[c]{8$\times$8} & \makecell[c]{16 depth levels}& \makecell[c]{24.8622}& \makecell[c]{0.8164}\\
        \hline
        \makecell[c]{8$\times$8} & \makecell[c]{32 depth levels}& \makecell[c]{\textbf{24.8917}}& \makecell[c]{\textbf{0.8174}}\\
        \hline
        \makecell[c]{8$\times$8} & \makecell[c]{48 depth levels}& \makecell[c]{24.8707}& \makecell[c]{0.8165}\\
        \hline

  \end{tabular}
}
\label{ablation}
\end{table}

\subsection{Ablation studies}
\label{section45}

\textbf{Contextual Correlation.}
The CCL can be implemented in three steps and we carry on ablation experiments on these steps. We validate the effectiveness of every stage using the baseline network (shown in Fig. \ref{baseline}) on Warped MS-COCO, and the results are shown in Table \ref{ablation_corraltion}.
``Feature Flow'' can convert correlation volume into a simple but efficient representation. And ``Scale Softmax'' can further improve the matching performance by effectively suppressing the interference of points with low matching probability on feature flow.

\begin{spacing}{1.5}
\end{spacing}
\textbf{Number of Grids.}
The increase of the grids' number can bring about an improvement in image alignment. But it is not absolute since the network architecture and the size of the dataset may affect it. Therefore, we explore the best number of grids in our work and the results are shown in Row 2-6 of Table \ref{ablation}. According to our experiment, $8\times 8$ mesh is the best.

\begin{figure}[!t]
    \centering
    \includegraphics[width=0.47\textwidth]{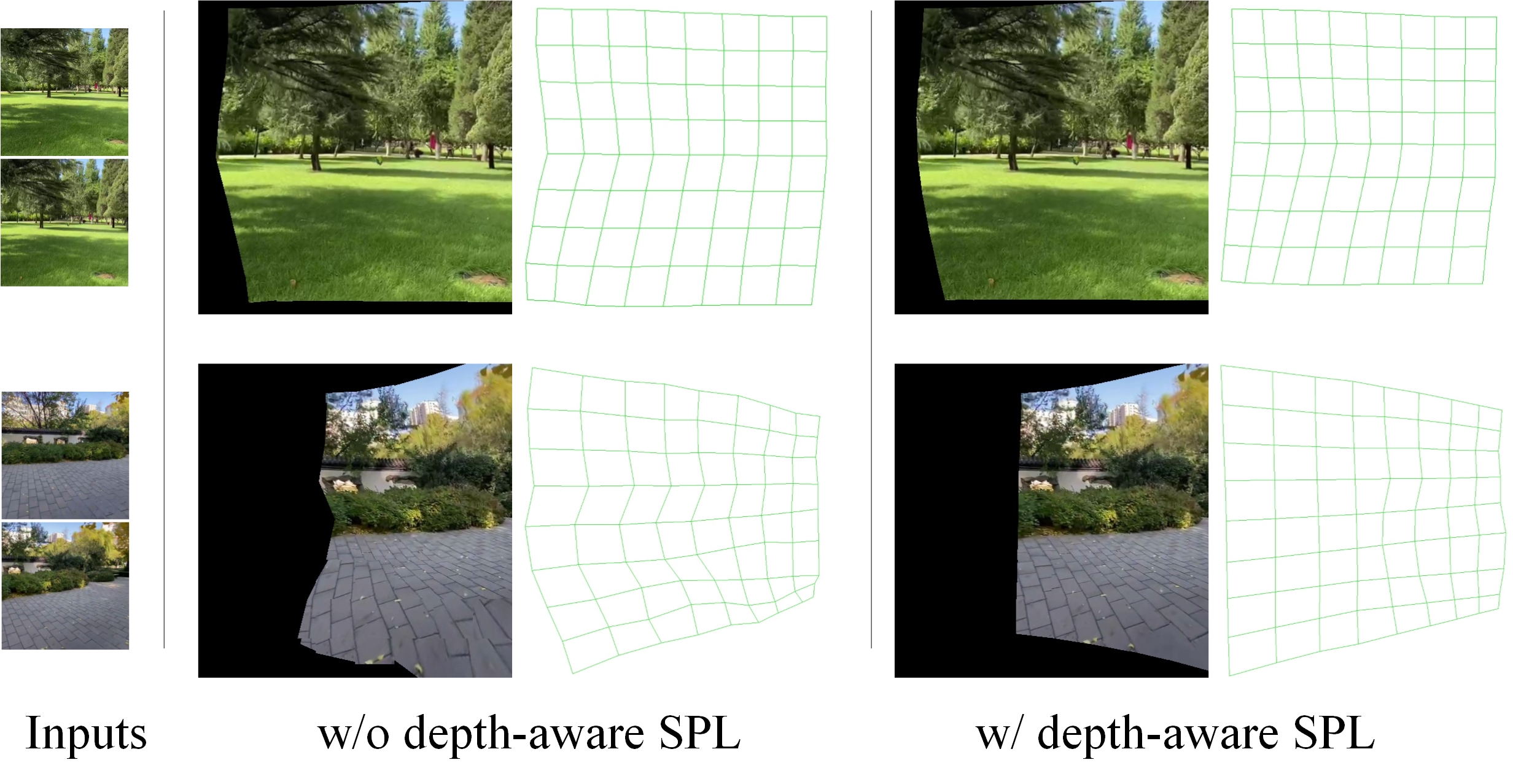}
    \vspace{-0.4cm}
    \caption{Qualitative comparison on the mesh shape. The warped target image and the deformated mesh are shown in pair.}
    \label{shape-preserved}
\end{figure}

\begin{spacing}{1.5}
\end{spacing}
\textbf{Depth-Aware Shape-Preserved Loss.}
The shape-preserved constraint can maintain the rigid shape of the mesh. But it regards all the grids as being on the same plane, reducing the alignment performance of the network in the existence of parallax. Dividing an image mesh into different depth levels can avoid this problem. We explore the best number of depth levels in our network, and the results are shown in Row 5, 7-13 of Table \ref{ablation}, where Row 5 can be regarded as no shape-preserved constraint and Row 7 can be considered as a global shape-preserved constraint (as \cite{wang2018deep, ye2019deepmeshflow} do). Also, a qualitative comparison of the mesh shape is provided in Fig. \ref{shape-preserved} to demonstrate the superiority of our depth-aware shape-preserved loss.

\section{Conclusion and Future Prospect}
\label{section5}
In this paper, we have proposed a depth-aware multi-grid deep homography estimation network to align images with parallax from global to local, breaking through the limitation of the existing single deep homography estimation solutions. In our network, we design the CCL to extract the matching relationship efficiently, outperforming the cost volume in performance, the number of parameters, and the speed. Besides, a depth-aware shape-preserved loss is presented to improve the shape regularity and alignment performance, simultaneously. Extensive experiments prove our superiority to the existing single homography and multi-homography solutions.

However, the number of grids can be limited by the network architecture and the size of the dataset. In the future, we would like to explore the reason that affects the number of grids and increase the upper limit of grids' number without decreasing alignment performance.

\normalem
\bibliographystyle{ieeetr}
\bibliography{reference}
\end{document}